%% file: main.tex
\documentclass[10pt,twocolumn,letterpaper]{article}

\usepackage[pagenumbers]{cvpr} 
\usepackage{algorithm}
\usepackage{algorithmic}
\usepackage{listings}
\DeclareCaptionStyle{ruled}{labelfont=normalfont,labelsep=colon,strut=off} 
\floatstyle{ruled}

\usepackage{newfloat}
\newfloat{listing}{tb}{lst}{}
\floatname{listing}{Listing}
\input{preamble}

\definecolor{cvprblue}{rgb}{0.21,0.49,0.74}
\usepackage[pagebackref,breaklinks,colorlinks,allcolors=cvprblue]{hyperref}
\usepackage{url}
 
\makeatletter
\renewcommand{\and}{\hspace{1em}}
\makeatother
\title{DreamVVT: Mastering Realistic Video Virtual Try-On in the Wild via a Stage-Wise Diffusion Transformer Framework}
\author{
    Tongchun Zuo\textsuperscript{\rm 1}$^\star$ 
    \and
    Zaiyu Huang\textsuperscript{\rm 1}$^\star$ 
    \and
    Shuliang Ning\textsuperscript{\rm 1} 
    \and
    Ente Lin \textsuperscript{\rm 2}     
    \and
    Chao Liang\textsuperscript{\rm 1} \\
    \and
    Zerong Zheng\textsuperscript{\rm 1}  
    \and
    Jianwen Jiang\textsuperscript{\rm 1} 
    \and
    Yuan Zhang\textsuperscript{\rm 1} 
    \and
    Mingyuan Gao\textsuperscript{\rm 1} 
    \and
    Xin Dong\textsuperscript{\rm 1}$^\dagger$ 
    \\
    \textsuperscript{\rm 1}ByteDance Intelligent Creation, 
    \textsuperscript{\rm 2}Shenzhen International Graduate School, Tsinghua University \\
    ztcustc@mail.ustc.edu.cn, {huangzaiyu,liangchao,dongxin.1016}@bytedance.com, \\ linet22@mails.tsinghua.edu.cn, shuliangning@link.cuhk.edu.cn, zrzheng1995@foxmail.com, \\{jianwen.alan,zhang.yuan09,gaomingyuan001}@gmail.com
}

\usepackage{bibentry}
\usepackage{enumitem}
 
\begin{document}

\twocolumn[{
\renewcommand\twocolumn[1][]{#1} 
    \maketitle
    \centering
    \vspace{-2.5em}
    \captionsetup{type=figure}
    \includegraphics[width=1\textwidth]{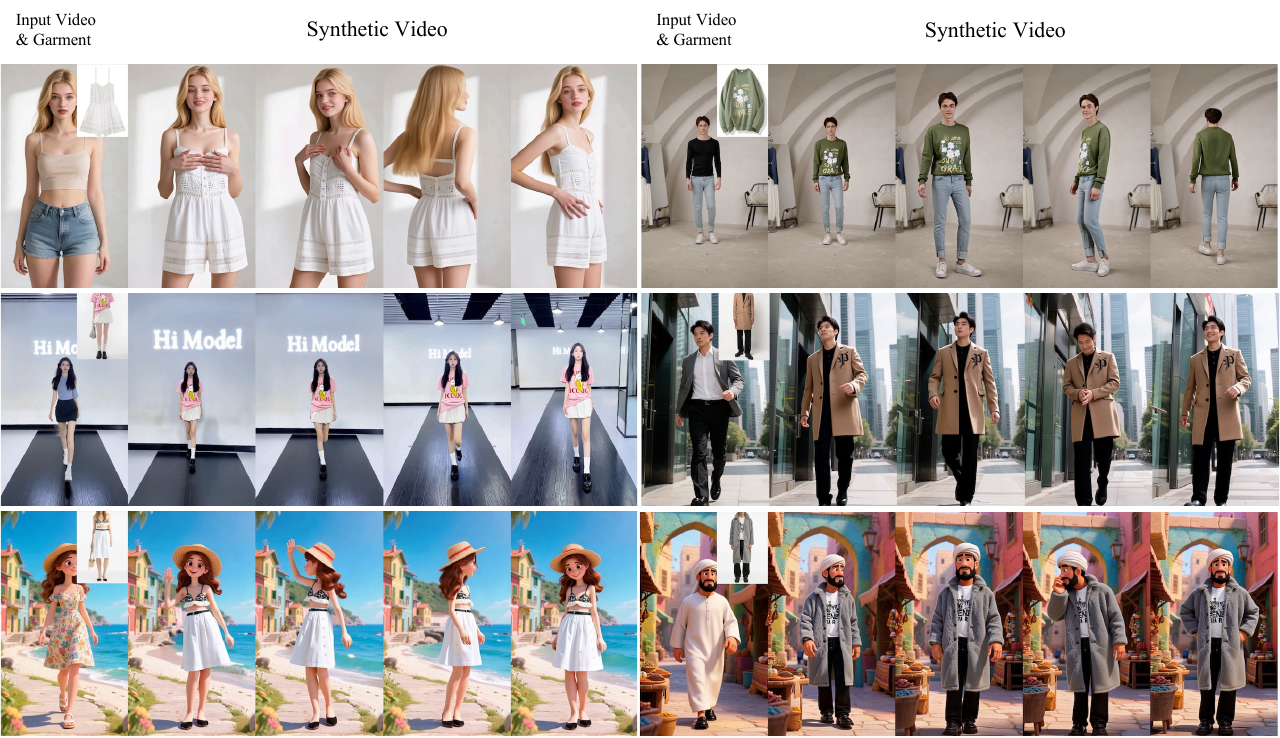}
    \captionof{figure}{DreamVVT can generate high-fidelity and temporally coherent virtual try-on videos for diverse garments and in unconstrained scenarios. Specifically, \textbf{the first row} shows its ability to handle complex human motions like runway walks and 360-degree rotations;\textbf{the second row} illustrates robustness to complex backgrounds and challenging camera movements; \textbf{the third row} highlights visually coherent try-on results for cartoon characters with real garments.}
    \label{fig:teaser}
}]

\renewcommand{\thefootnote}{}
\footnotetext{$^\star$ Equal contribution.}
\footnotetext{$^\dagger$ Project leader.}


\input{sec/abstract}
\input{sec/introduction}

\input{sec/related_work}

\input{sec/method}

\input{sec/experiments}
\input{sec/conclusion}
\input{sec/Acknowledgement}
 
{
    \small
    \bibliographystyle{ieeenat_fullname}
    \bibliography{main}
}

\input{sec/supple_for_arxiv}

\end{document}

%% file: preamble.tex
\usepackage[utf8]{inputenc}


%% file: sec/abstract.tex
\begin{abstract}
Video virtual try-on (VVT) technology has garnered considerable academic interest owing to its promising applications in e-commerce advertising and entertainment. However, most existing end-to-end methods rely heavily on scarce paired garment-centric datasets and fail to effectively leverage priors of advanced visual models  and test-time inputs, making it challenging to accurately preserve fine-grained garment details and maintain temporal consistency in unconstrained scenarios.
To address these challenges, we propose \textbf{DreamVVT}, a carefully designed two-stage framework built upon Diffusion Transformers (DiTs), which is inherently capable of leveraging diverse unpaired human-centric data to enhance adaptability in real-world scenarios. 
To further leverage prior knowledge from  pretrained models and  test-time inputs, in the first stage, we sample representative frames from the input video and utilize a multi-frame try-on model integrated with a vision-language model (VLM), to synthesize high-fidelity and semantically consistent keyframe try-on images. These images serve as complementary appearance guidance for subsequent video generation.
\textbf{In the second stage},  skeleton maps together with fine-grained motion and appearance descriptions are extracted from the input content, and these along with the keyframe try-on images are then fed into a pretrained video generation model enhanced with LoRA adapters. This ensures long-term temporal coherence for unseen regions and enables highly plausible dynamic motions.
Extensive quantitative and qualitative experiments demonstrate that DreamVVT surpasses existing methods in preserving detailed garment content and temporal stability in real-world scenarios. Our project page \url{https://virtu-lab.github.io/}
 
\end{abstract}
 

%% file: sec/introduction.tex
\enlargethispage{1\baselineskip}
\section{Introduction}
With the scaling up of training data and model parameters, Diffusion Transformer (DiT)-based models have achieved remarkable progress in various visual generation tasks, including text-to-image \cite{esser2024scaling,flux2024} and image-to-video generation \cite{wan2025,li2024hunyuandit}.
As a prominent downstream task, video virtual try-on (VVT) aims to faithfully  render arbitrary garments onto characters within video sequences, as shown in Figure~\ref{fig:teaser}. Recently, it has attracted considerable attention from the research community \cite{vace, zheng2024dynamic, chong2025catv2ton, li2025magictryon} owing to its broad application potential in promising domains such as e-commerce and entertainment. 

Despite considerable efforts, existing methods\cite{zheng2024dynamic, chong2025catv2ton, li2025magictryon, xu2024tunnel, blattmann2023stable, fang2024vivid, li2025pursuing, li2025realvvt} still struggle to accurately preserve fine-grained garment details and maintain temporal consistency in unconstrained scenarios, such as complex subject or camera motion, dynamic scenes, and diverse character styles.  We posit that these limitations stem primarily from the reliance on an end-to-end training paradigm, which inherently constrains the effective exploitation of unpaired data, priors of advanced visual models, and additional information  at inference stage. 
\textbf{First},  these methods exhibit a strong reliance on insufficient paired clothing-video data \cite{dong2019fw,fang2024vivid}, most of which are collected in homogeneous indoor environments. This often results in reduced garment visual fidelity and increased temporal instability, particularly for arbitrary garments and complex video inputs. Furthermore, collecting large-scale paired clothing–video datasets across diverse real-world scenarios remains extremely challenging.  
\textbf{Second}, these methods\cite{zheng2024dynamic, chong2025catv2ton, li2025magictryon} adapt a pretrained text-to-video generation model to deform spatially misaligned garment images onto the person frame by frame. However, this approach disrupts the pretrained model’s inherent capacity for smooth spatiotemporal modeling, making model convergence more challenging.
Additionally, performing full fine-tuning of all parameters \cite{blattmann2023stable,zheng2024dynamic,li2025magictryon} in a pretrained model with a limited amount of data is prone to disrupting the pretrained priors, which in turn degrades the quality and temporal stability of the generated videos. Even when a substantial portion of model parameters are trained on large-scale datasets and diverse video tasks, unified video creation and editing methods \cite{ye2025unic, vace, hu2025hunyuancustom} still struggle to accurately preserve garment details and maintain temporal coherence, primarily due to the lack of task-specific design for virtual try-on.
\textbf{Third},  at the inference stage, providing only front-view garment images to guide the virtual try-on process often leads to implausible outcomes for invisible regions when the person turns or the camera viewpoint changes significantly. 


To tackle these issues, we introduce \textbf{DreamVVT}, an improved stage-wise framework built upon Diffusion Transformers (DiTs), which is inherently capable of leveraging unpaired human-centric data from diverse sources to improve generalization in real-world scenarios. To further exploit prior information from pretrained models and the inference process,  in the first stage, we first sample keyframes with significant motion changes from input video, a vision-language model (VLM) is then employed to generate textual descriptions that map the input garment to each keyframe. These descriptions, along with garment images and other relevant conditions, are provided to a multi-frame try-on model equipped with LoRA \cite{hu2022lora} adapters, resulting in high-fidelity and semantically consistent try-on images for each keyframe. These images serve as  complementary appearance guidance for subsequent video generation.
In the second stage, we employ a temporally smooth pose guider for skeletal feature encoding and utilize an advanced video large language model (video LLM) to extract fine-grained action descriptions and other high-level visual information from input content. These features, together with the spatially aligned keyframe try-on images, are subsequently provided as inputs to a pretrained video generation model enhanced with LoRA adapters. By leveraging the pretrained priors of large-scale video generation models, this model is endowed with enhanced generalization capability in in-the-wild scenarios. Moreover, incorporating multiple keyframe try-on images and precise motion guidance enables the generation of long-term virtual try-on videos that exhibit strong temporal consistency and highly plausible dynamic motions. Additionally, a multi-task learning strategy is introduced to maintain the controllability of conditions from different modalities.

In summary, the contributions of this paper are threefold:
\begin{itemize}
    \item We present a carefully designed  stage-wise framework based on DiTs, which is capable of leveraging unpaired data, priors from advanced visual models  and test-time inputs to enhance virtual try-on performance in real-world scenarios.
    
    \item We propose integrating keyframe try-on with video LLM reasoning, which supplies abundant appearance and motion information for video generation and ensures both garment detail preservation and temporal consistency.
     
    \item Extensive experiments demonstrate that DreamVVT outperforms current existing methods  in preserving high-fidelity garment details and ensuring temporal stability across diverse scenarios.
\end{itemize}

%% file: sec/related_work.tex
\enlargethispage{1\baselineskip}

\begin{figure*}[t]
  \centering
    \captionsetup{type=figure}
    \includegraphics[width=1\textwidth]{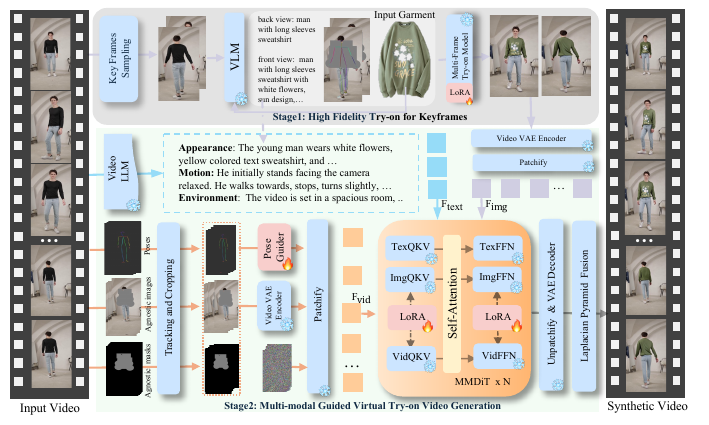}
  \caption{Overview of DreamVVT. The framework comprises two sequential stages: the first stage selects frames with significant motion changes and generates try-on images for these keyframes, while the second stage synthesizes the final virtual try-on video using fine-grained motion guidance and complementary appearance cues.} 
  \label{fig:framework}
\vspace{-5mm}
\end{figure*}

\section{Related work}
\subsection{Image-based Virtual Try-on}
A series of works\cite{morelli2023ladi,gou2023taming,xu2025ootdiffusion,kim2024stableviton,lin2025dreamfit,wang2025mv,guo2025any2anytryon,zhang2024boow} that build on powerful SDXL\cite{podell2023sdxl} and Flux\cite{flux2024}, achieve a significant improvement in realism of the synthesis results compared to previous GAN-based methods\cite{choi2021viton,xie2023gp}.
However, these methods often failed to preserve the fine-grained detail of the garment image because they heavily relies on the garment feature extracted by a pre-trained image clip encoder. To address these problems, some other pioneering approaches\cite{xu2025ootdiffusion,sun2024outfitanyone} proposed to introduce a parallel reference U-net architecture to effectively extract the reference tokens and then merge with the original tokens through the attention process. 
Dreamfit\cite{lin2025dreamfit} introduces a plug-and-play fine-tuning mechanism by leveraging LoRA\cite{hu2022lora} and parallel architecture.
While CatVTON\cite{chong2024catvton} presents a light-weighted approach, which replaces the parallel architecture with a shared-weighted single one and directly injects the conditional image through input concatenation. 
Although these methods demonstrate a superior performance in garment preservation, most of them are limited in conventional single-view virtual try-on settings without fully exploring the multi-images generation ability of large diffusion model.
Our method takes a step forward focus on instruction-guided key frame try-on image generation, which expands the border of try-on applications and further facilitates the usage of generated results to support downstream tasks like video try-on.

\subsection{Video-based Virtual Try-on} 

In recent works,  \cite{xu2024tunnel,fang2024vivid,  li2025pursuing}  employ a pretrained inpainting U-Net as the primary branch, complemented by a reference U-Net to capture fine-grained garment features. Temporal consistency is further enhanced by introducing standard temporal attention modules after each stage of the main U-Net. Owing to their intrinsically decoupled spatiotemporal architecture, these approaches persistently exhibit pronounced temporal instability in try-on videos featuring complex garments.  To alleviate these issues, \cite{li2025realvvt} encompasses a Clothing \& Temporal Consistency strategy based on Stable Video Diffusion \cite{blattmann2023stable}.  \cite{zheng2024dynamic,chong2025catv2ton,li2025magictryon} adopt the Diffusion Transformer \cite{peebles2023scalable} backbone with full attention mechanisms to enhance temporal consistency, and further propose more efficient training strategies. Despite their promise, these single-stage methods are limited by their inability to leverage unpaired videos or images across a wide range of scenarios and their tendency to disrupt pretrained priors. Consequently, they struggle to simultaneously maintain visual fidelity and motion consistency, particularly when the input videos involve camera movements, dynamic backgrounds, or complex human actions. In this work, we  divide the VVT task into two sequential stages, enabling efficient use of unpaired videos and images from diverse sources and fully exploiting the priors of pretrained models, thereby achieving significant improvements in unconstrained scenarios.


\subsection{Pose guided Human Video generation}
GAN-based methods\cite{liu2019liquid,ren2020deep,ren2022neural} and diffusion-based approaches\cite{zhang2024mimicmotion,wang2024unianimate,hu2024animate,hu2025animate,men2024mimo} form the main-stream implementation paradigm of human video generation in recent years, which mostly contains an individual reference encoder responsible for extracting image features and a pose guider for injecting motion signal.
In particular, AnimateAnyone\cite{hu2024animate} first introduces the popular parallel u-net architecture based on the stable diffusion model. Mimo\cite{men2024mimo} subsequently proposes to decouple the background and foreground signal by adopting an expert encoder for different conditional inputs. AnimateAnyone2\cite{hu2025animate} focus on character-environment integration and pose robustness through a new environment and pose formulation strategy. 
We observe that methods using an entire image as a reference frequently degrade garment details, especially when the subject or camera undergoes large rotations. Therefore, to achieve high-fidelity video try-on, we develop a video editing model guided by multiple try-on keyframes, incorporating human tracking-based cropping and multi-task learning.

%% file: sec/method.tex

\section{Methodology}
Our DreamVVT adopts a stage-wise framework based on large-scale Diffusion Transformers to achieve high-fidelity virtual try-on video generation in unconstrained scenarios. As illustrated in Figure \ref{fig:framework}, it consists of two sequential stages. In the first stage, we sample frames with notable motion variations from a input video as keyframes, and then a multi-frame try-on model is developed to fit garment images onto these keyframes while maintaining content consistency and preserving fine details. In the second stage, we present a modified video generation model, which synthesizes a plausible try-on video conditioned on keyframe try-on images, pose features, and textual descriptions. We further elaborate on each specific module within our framework, emphasizing its importance and role in the overall framework.

\subsection{Input conditions}
\textbf{Pose Conditions.} To enable virtual try-on for both real and cartoon characters, we adopt RTMPose \cite{jiang2023rtmpose} as a robust and efficient pose representation. Given that the character in the raw video may occupies only a limited spatial area, directly feeding the pose sequence at its original resolution into the try-on model can result in significant garment detail loss in the generated video, primarily due to spatial downsampling. To address this issue, We begin by cropping each frame with tracking bounding boxes of uniform width and height, thereby isolating the character region in each frame. \\
\textbf{Agnostic Masks.} 
Most previous methods \cite{chong2025catv2ton,li2025magictryon,xu2024tunnel}  generate agnostic masks by directly dilating the segmented clothing regions in the video, which can easily cause leakage of the original garment style cues. To mitigate this problem, we employ the human bounding box together with the dilated pose skeleton to generate clothing-agnostic masks that effectively prevent information leakage while preserving as much of the original background as possible. \\
\textbf{Agnostic Images.}
By applying the  agnostic masks, we occlude the garment regions in the input character video or image, thereby generating  agnostic images. \\
\textbf{Garment Images.} 
For garment image input, we firstly employ a saliency segmentation detection model\cite{zheng2024birefnet} to extract the foreground area and then remove the background region by filling it with white pixels. To further promote the preservation of garment details, we calculate a tight bounding box according to the extracted segmentation and subsequently crop the region of interest. Finally, the cropped image is resized to a specific resolution before being fed into the network.  

\begin{figure*}[t]
  \centering
    \captionsetup{type=figure}
    \includegraphics[width=1\textwidth]{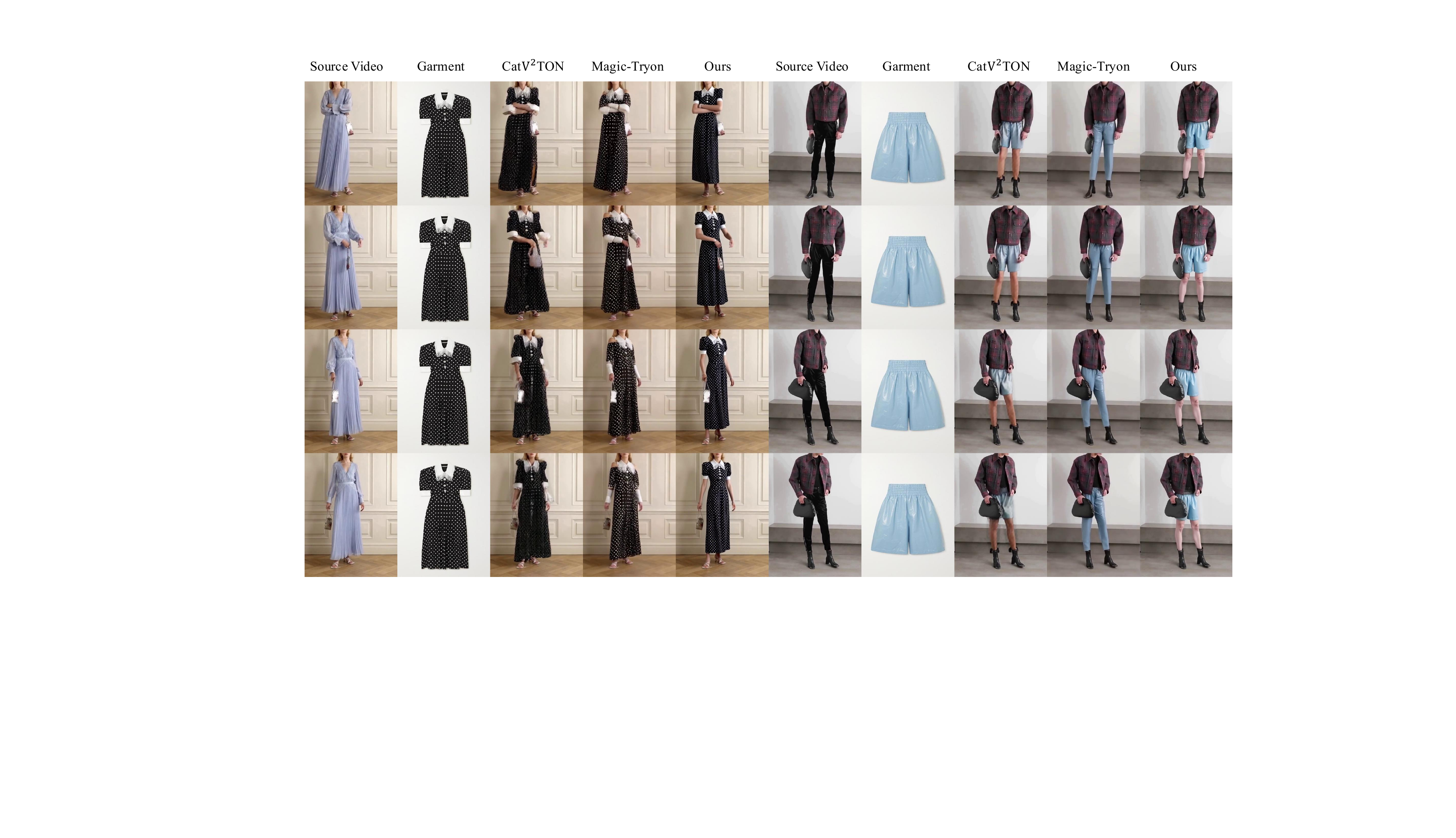}
  \caption{Qualitative comparison on the ViViD dataset.Please zoom in for more details.} 
  \label{fig:qualitative_vivid}
\vspace{-5mm}
\end{figure*}

\subsection{Stage1: High Fidelity Try-on for Keyframes} 
\subsubsection{Keyframe Sampling}

We select frames with significant motion changes to provide more comprehensive guidance for video generation.
Initially, given that the majority of input garment images are captured from a frontal perspective, we predefine a frontal-view person image in an A-pose as the anchor frame. 
Subsequently, we compute the motion similarity between each video frame and the anchor frame by measuring the cosine distance between their respective skeletal joint direction vectors. This similarity is further weighted by the area ratio of the subject to the entire frame, to produce the final score.
Lastly, frames are sorted in descending order by their final scores, and a reverse-order search constrained by a minimum score interval is performed to obtain a set of key images with minimal informational redundancy. The  details are provided in the Supplementary Material.

\subsubsection{Multi-frame Try-on Model }
Given selected keyframes, we leverage a diffusion transformer $\mathcal{G}^{*}$ featuring a minimal set of learnable parameters on a pre-trained Seedream\cite{gao2025seedream} model $\mathcal{G}$ to generate the final multi-frames try-on results. We modify each MMDiT blocks\cite{esser2024scaling} within $\mathcal{G}$ by integrating the attention modules with pluggable LoRA\cite{hu2022lora} layers, and introduce an additional parameter-sharing network branch to process reference image input following implementations of \cite{lin2025dreamfit}. Notably, $\mathcal{G}^{*}$ takes as input the multi-keyframes image-condition pairs alongside an elaborately designed consistent image instruction, which clarifies different input components and helps guide the model towards synthesizing desirable outcomes. Specifically, we first tokenize each conditional input via a parallel network architecture to align different modalities, then jointly aggregate information across keyframes through $Q$, $K$, $V$ catenation during the attention process. This mechanism ensures robust information interaction between each conditional input and keyframes intermediate features, thereby enabling the coherent multi-frames try-on results with consistent details. For text inputs, we resort to \cite{seed2025seed1_5vl} for detailed descriptions, including garment category, material, and patterns for each keyframe. Subsequently, a text alignment procedure is introduced by asking VLM to rewrite and gather all the text results, which further reinforces the consistency of keyframe descriptions.

\begin{figure*}[t]
  \centering
    \captionsetup{type=figure}
    \includegraphics[width=1\textwidth]{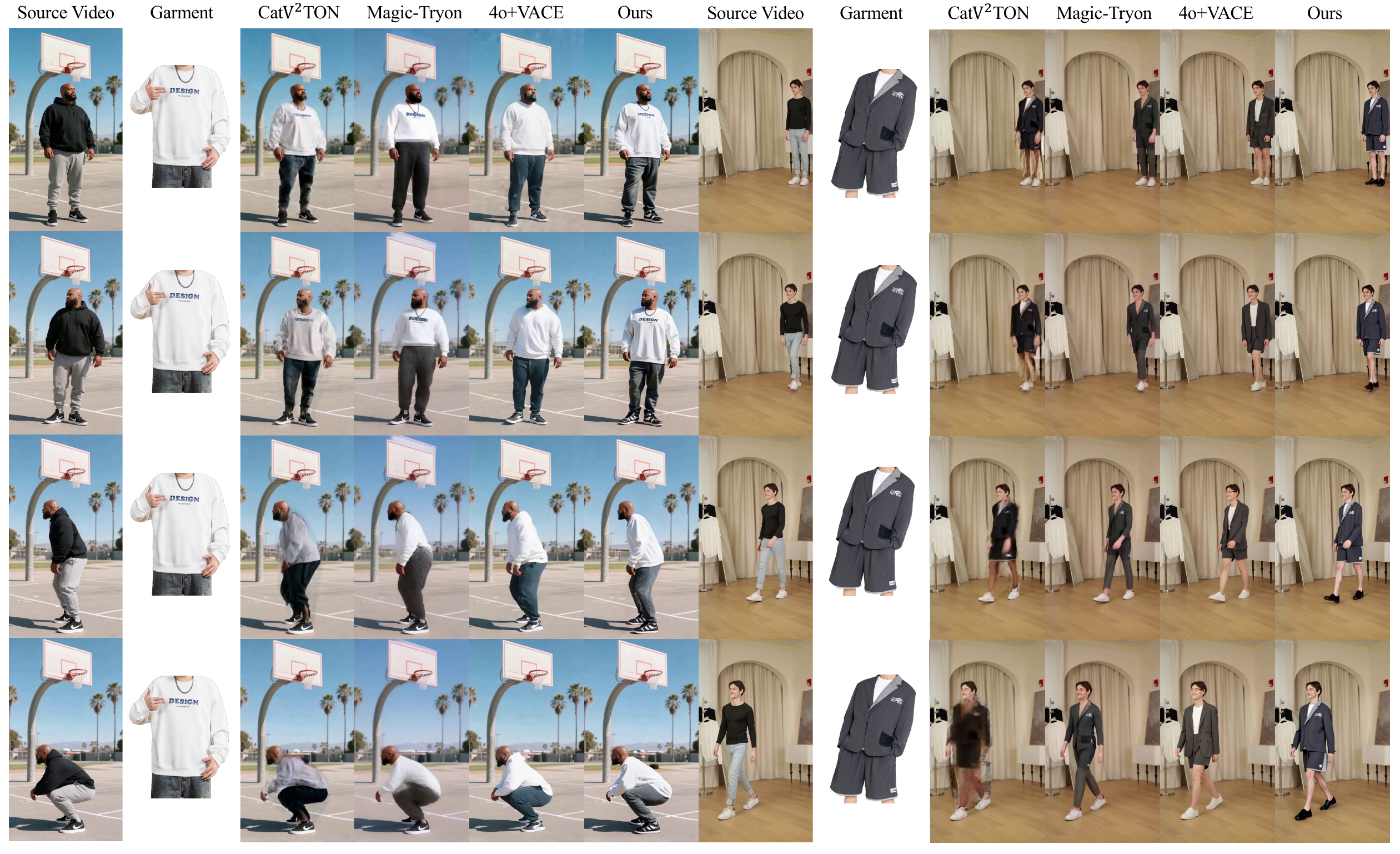}
  \caption{Qualitative results on our Wild-TryOn Benchmark. Please zoom in for more detail. Additional comparison results are provided in the supplementary material.} 
  \label{fig:qualitative_ours}
\vspace{-5mm}
\end{figure*}

\subsection{Stage2: Multi-modal Guided Virtual Try-on Video Generation  }
Our virtual video try-on model is based on a pretrained image-to-video generation framework \cite{lin2025diffusion} that employs stacked sequential MMDiT blocks \cite{esser2024scaling}, each of which integrates text and video streams. To accurately reconstruct body movements in the input video, we extract the corresponding 2D skeleton sequences. 
After cropping, a tailored pose guider with temporal attention transforms the frame-wise skeleton maps into temporally smoothed pose latents that match the resolution of the noise latents. Similarly, the cropped agnostic images are fed into the video VAE encoder to obtain agnostic latents, and the cropped agnostic mask is resized to  the same resolution as the agnostic latents. Subsequently, the agnostic latents, resized agnostic masks, noise latents, and pose latents are concatenated along the channel dimension and patchified into video tokens, denoted as \(F_{\text{vid}}  \in \mathbb{R}^{l_v \times c} \) (where $l_v = t \times h \times w$, and $t = \frac{T}{4}, h = \frac{H}{16}, w = \frac{W}{16}$, $T, H, W$ is the shape of input video)
Furthermore, since pose skeletons capture only coarse-grained body motion and cannot fully represent fine-grained garment interactions, we employ Qwen2.5-VL\cite{Qwen2.5-VL} to extract an attribute-disentangled textual description, which contains detailed motion descriptions and high-level visual information(During inference, the appearance-related descriptions are replaced with those corresponding to the target garment). These textual descriptions are then processed by a Qwen LLM \cite{qwen} into text tokens, denoted as \(F_{\text{text}} \in \mathbb{R}^{l_t \times c_t}\).
For the appearance branch, keyframe try-on images are first processed frame by frame by the video VAE encoder to extract image latents, which are then transformed into image tokens, denoted as \(F_{\text{img}} \in \mathbb{R}^{l_i \times c}\) (where $l_i = k \times h \times w$ and k is the number of keyframes). 
To preserve the spatiotemporal modeling and prompt adherence capabilities of the model, we freeze the parameters of the text streams. Lightweight LoRA adapters, comprising only 10\% of the trainable parameters, are inserted into the frozen video streams and image streams that are directly duplicated from the video streams, with shared memory. As the channel dimensions of the video and image tokens have increased, the input projection layers for the video and image streams are set to be trainable.
Finally, all these token sets are processed by their respective QKV projection layers and then concatenated along the \(l\) dimension. The resulting sequence is fed into a full self-attention module, which enables the model to adaptively align visual content with textual descriptions across both spatial and temporal dimensions. Following the self-attention operation, the joint tokens are demultiplexed by index into text, image, and video tokens, which are subsequently processed by the following DiT blocks.
After several denoising iterations within the DiT backbone, the network generates the try-on video tokens, which are then decoded into video sequences by the Video VAE decoder. An efficient laplacian pyramid fusion method is subsequently applied to seamlessly blend the generated try-on video into the corresponding regions of the original video.
During training, we introduce a multi-task learning strategy similar to \cite{lin2025omnihuman1,jiang2025mobileportrait}, wherein one task (e.g., text to video,  pose with text and keyframes to video) is randomly selected based on a predefined probabilistic schedule, in order to fully exploit the complementary advantages of various modalities.  


%% file: sec/experiments.tex
\begin{figure*}[t]
  \centering
    \captionsetup{type=figure}
    \includegraphics[width=1\textwidth]{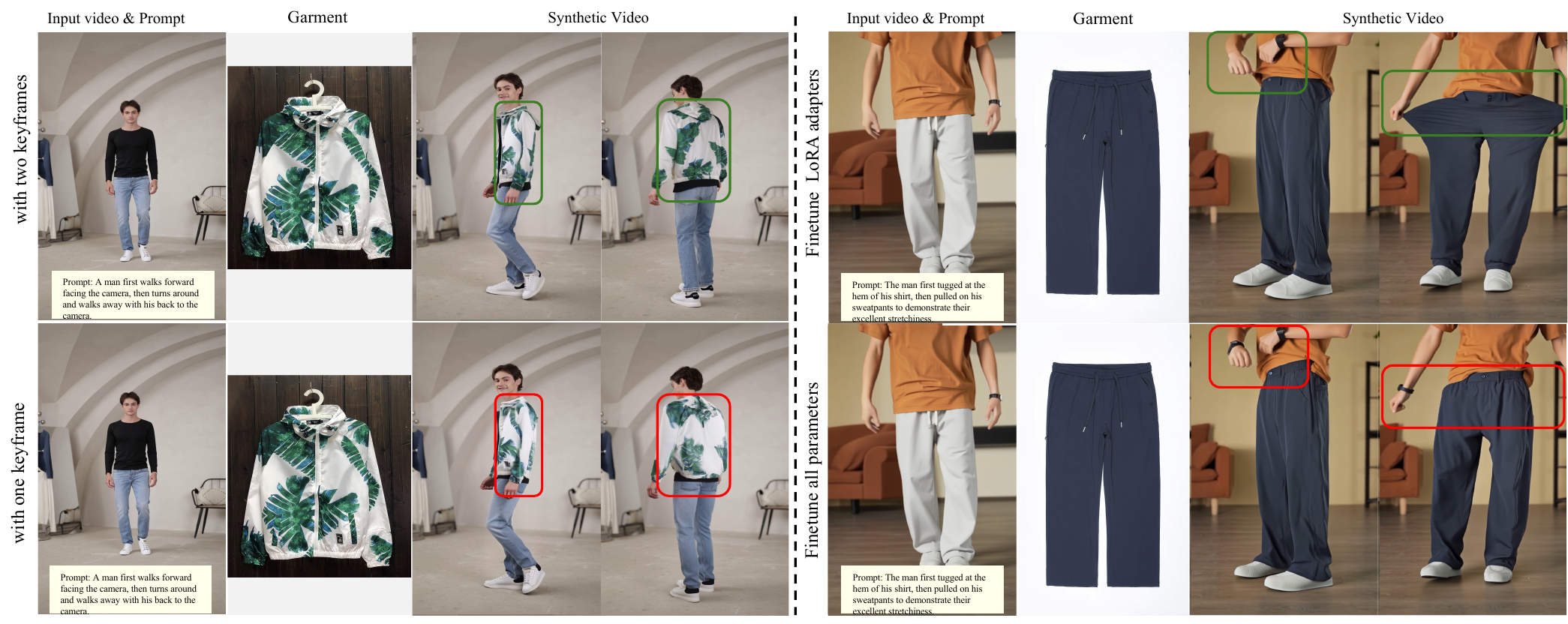}
  \caption{\textbf{Left}: Ablation studies with different keyframes. \textbf{Right}: Ablation studies with finetune all parameters or LoRA adapters} 
  \label{fig:abalation_frame_lora}
\vspace{-5mm}
\end{figure*}

\begin{table}[t]
\small
\def\arraystretch{1.0}
\tabcolsep 0pt
\centering
    \begin{tabular}{c c cccccc}
      \toprule
      Dataset & & \multicolumn{6}{c}{ViViD} \\
      \cmidrule{1-8} 
      Method & & VFID$^p_I$ $\downarrow$ & VFID$^p_R$ $\downarrow$ & 
      VFID$^u_I$ $\downarrow$ & VFID$^u_R$ $\downarrow$ &
      SSIM$\uparrow$ & LPIPS$\downarrow$\\
      \cmidrule{1-8} 
      ViViD & & 17.2924 & 0.6209 & 21.8032 & 0.8212 & 0.8029 & 0.1221 \\
      CatV$^2$TON & & 13.5962 & 0.2963 & 19.5131 & 0.5283 & 0.8727 & 0.0639\\
      MagicTryON & & 12.1988 & \textbf{0.2346} & 17.5710 & 0.5073 & \textbf{0.8841} & 0.0815 \\
      DreamVVT & & \textbf{11.0180} & 0.2549 & \textbf{16.9468} & \textbf{0.4285} & 0.8737 & \textbf{0.0619} \\
      \bottomrule
    \end{tabular}
    \caption{Quantitative comparisons on ViViD dataset. The best results are demonstrated on \textbf{bold}.}
\vspace{-5mm}
\label{tab:quantitative-vivid}
\end{table}

\section{Experiments}
\subsection{Datasets}
We curated a high-quality human-centric video dataset comprising 69,643 samples, characterized by unrestricted subject and camera movements as well as dynamic scenes. Additionally, over one million pairs of  multi-view images of the same individual were gathered from public websites. We performed mixed training using the collected unpaired data in combination with three publicly available try-on datasets: VITON-HD \cite{choi2021viton}, DressCode \cite{morelli2022dress}, and ViViD \cite{fang2024vivid}.
 During testing, we conducted indoor scenario evaluations on the ViViD-S dataset, which contains 180 samples, following the methodology of \cite{chong2025catv2ton}.
Additionally,  we created an in-the-wild benchmark, named Wild-TryOnBench, consisting of  81 samples that encompass rich variations in subject or camera movement, scene changes, diverse forms of garment input and character styles.

\begin{table}[t]
\small
\def\arraystretch{1.0}
\tabcolsep 8pt
\centering
\begin{tabular}{lccc}
  \toprule
  Method & GP $\uparrow$ & PR $\uparrow$ & TC $\uparrow$ \\
  \midrule
  CatV$^2$TON\cite{chong2025catv2ton} & 1.30 & 1.04 & 1.08 \\
  MagicTryON\cite{li2025magictryon} & 1.19 & 1.81 & 1.88 \\
  GPT4o+VACE\cite{vace} & 2.67 & 3.51 & 2.61 \\
  DreamVVT & \textbf{3.41} & \textbf{3.69} & \textbf{3.32} \\
  \bottomrule
\end{tabular}
\caption{Quantitative comparisons on the Wild-TryOnBench dataset. We use GP, PR, TC as the short for Garment Preservation, Physical Realism and Temporal Consistency.}
\vspace{-5mm}
\label{tab:quantitative_comparison}
\end{table}

\subsection{Implementation Details}
We used the AdamW optimizer with a constant learning rate of 2e-5, weight decay set to 0.01, and gradient clipping set to 1.0. The entire training process was conducted on 8 NVIDIA H20(96G) GPUs for about 10 days. The LoRA rank of the model is set to 256 and the model is trained using the standard generation loss of rectified flow \cite{esser2024scaling}. During inference, we employ the Euler scheduler \cite{2022Elucidating} with 50 sampling steps, set the classifier-free guidance scale to 2.5, and fix the random seed at 42. For keyframe try-on, each result is generated three times, and the best outcome is selected. In contrast, for video generation, each result is produced in a single run.  Please refer to the  supplementary material for details regarding the video and image caption, as well as the implementation of training and inference procedures for long videos. 

\subsubsection{Evaluation}
The evaluation metrics follow those adopted in the previous works, whereas VFID with I3D\cite{carreira2017quo}and ResNext\cite{xie2017aggregated} is introduced to evaluate video quality in unpaired scenarios, while SSIM\cite{Wang2004SSIM}, LPIPS\cite{zhang2018unreasonable} is added to measure the similarity between synthesized results and ground-truth in paired try-on test.
Note that the VFID metric is mainly proposed to measure the quality of the video. 
We further introduce human evaluation on our Wild-TryOn benchmark, which incorporates three evaluation criteria: garment detail preservation, physical realism, and temporal consistency. Each criterion is scored on a scale of 0 to 5, with 0 indicating the worst performance and 5 the best.

\subsection{Qualitative Comparison}
For qualitative evaluation, we compare our method with advanced video virtual try-on methods including CatV$^2$TON, MagicTryOn and GPT4o+VACE. As shown in Fig.\ref{fig:qualitative_vivid} and Fig.\ref{fig:qualitative_ours}, our method demonstrates a strong superiority in synthesizing realistic and spatiotemporally smooth virtual try-on results in both the ViViD-S dataset and our Wild-TryOn evaluation benchmark. In contrast, CatV$^2$TON\cite{chong2025catv2ton} and MagicTryOn\cite{li2025magictryon} show limited scalability in handling in-the-wild scenarios due to scarce garment-video pairs for training, which results in drastic performance degradation in out-of-domain test cases in our Wild-TryOn benchmark. Note that CatV$^2$TON and MagicTryOn are prone to generate blurry results when encountering complex try-on scenarios such as 360-degree rotation, which often involves severe self-occlusion. GPT4o+VACE struggle to preserve the person's identity and reproduce cloth details from given images. Our method, however, outperforms previous methods in generalizabilty with robust synthesis quality on arbitrary resolution, frame rate, indoor and outdoor scenarios. 
\subsection{Quantitative Comparison}
For quantitative evaluation, as reported in Tab.\ref{tab:quantitative-vivid}, our method achieves state-of-the-art performance among compared baselines in both ViViD-S and our Wild-TryOn benchmark. Specifically, in the ViViD-S dataset, our method exceeds other baselines by a considerably large margin in unpaired try-on setting, with the lowest VFID score with I3D and ResNext. For paired try-on setting, our method achieves competitive results compared to existing video virtual try-on approaches, with the best performance on VID$_I$ and LPIPS metrics, and the second-best performance on VID$_R$ and SSIM metrics. Tab.\ref{tab:quantitative_comparison} shows similar observations, where our method outperforms existing baseline approaches, achieving the best performance on all three metrics.

\begin{table}[t]
\def\arraystretch{1.0}
\tabcolsep 8pt
\centering
\begin{tabular}{lccc}
  \toprule
  Method & GP $\uparrow$ & PR $\uparrow$ & TC $\uparrow$ \\
  \midrule
  DreamVVT-K1-w/o LoRA & 3.10 & 3.43 & 3.26  \\ 
  DreamVVT-K1-w LoRA & 3.16 & 3.62 & 3.29 \\ 
  DreamVVT-K2-w LoRA & \textbf{3.41} & \textbf{3.69} & \textbf{3.32} \\
  \bottomrule
\end{tabular}
\caption{Quantitative results for ablation studies on WildTryOn benchmark. K1 and K2 denote using one and two keyframes, respectively, while "w/o LoRA" refers to full-parameter fine-tuning.}
\vspace{-5mm}
\label{tab:quantitative_abalation}
\end{table}

\subsection{Ablation Study}
To validate the contribution of the proposed modifications to the final improvement, we further conduct ablation experiments in our Wild-TryOn benchmark. Specifically, we implement three variants of the model by training model with and without LoRA, and testing model with different numbers of keyframes. The human evaluation in Tab.\ref{tab:quantitative_abalation} is introduced to evaluate the performance of model with different components.
\subsubsection{Key frame number}
In the left panel of Fig.\ref{fig:abalation_frame_lora} show qualitative results of our DreamVVT with different numbers of keyframes. As demonstrated in the figure, integrating multiple number of keyframes helps enhance the model performance under complicated scenarios like turn around. Model with only one keyframe input is prone to produce blurry results or generate artifacts due to lack of garment detail information. In contrast, model with two keyframes produces much more plausible results with the clear patterns. The quantitative results further support this observation. The variant with two keyframes achieves an obvious improvement compared to one keyframe setting on detail preservation metric. Besides, with sufficient garment and motion information provided by two keyframes, the score of physical realism and temporal consistency also have a slight increase.
\subsubsection{LoRA adapters for video generation model}
As shown in the right panel of Fig.~\ref{fig:abalation_frame_lora}, when provided with the same textual description, the model trained with the LoRA adapter better preserves the pretrained model’s text control capability compared to full-parameter fine-tuning, thereby generating physically realistic clothing-interaction videos. The quantitative results in \ref{tab:quantitative_abalation} further demonstrate that this approach significantly improves the score of physical realism without compromising garment detail preservation or temporal consistency.

%% file: sec/conclusion.tex
\section{Discussions}
In this work, we present DreamVVT, a stagewise framework based on Diffusion Transformers (DiTs) that effectively leverages unpaired human-centric data, pretrained model priors, and test-time inputs by integrating keyframes try-on and multi-modal guided virtual try-on video generation. Extensive experiments demonstrate that DreamVVT surpasses state-of-the-art methods in preserving garment details and temporal consistency under unrestricted scenarios, and effectively handles diverse garments, highlighting its potential for e-commerce and entertainment applications.  

\textbf{Limitations.}
DreamVVT has several limitations. To accommodate arbitrary garment styles, the currently precomputed agnostic masks tend to cover large regions, which may compromise the integrity of both foreground objects and complex scenes. These challenges will be addressed in future research by employing mask-free video try-on techniques, for which DreamVVT will be utilized to construct the corresponding dataset. In addition, the method still struggles to achieve a high success rate in handling complex garment interaction motions, primarily due to limitations in the generative capabilities of pretrained models and the captioning of fine-grained actions. We plan to address and optimize this limitation in future work.

%% file: sec/Acknowledgement.tex
\section{Acknowledgement}
 We extend our sincere gratitude to Lu Jiang, Jiaqi Yang, Yanbo Zheng, Haoteng He, Yue Liu, Juan Li  for their invaluable contributions and supports to this research work.

%% file: sec/supple_for_arxiv.tex
\twocolumn[{
\renewcommand\twocolumn[1][]{#1}%
    \centering
    
  \section{Supplementary Material}
  \vspace{2em}
    \captionsetup{type=figure}
    \includegraphics[width=1\textwidth]{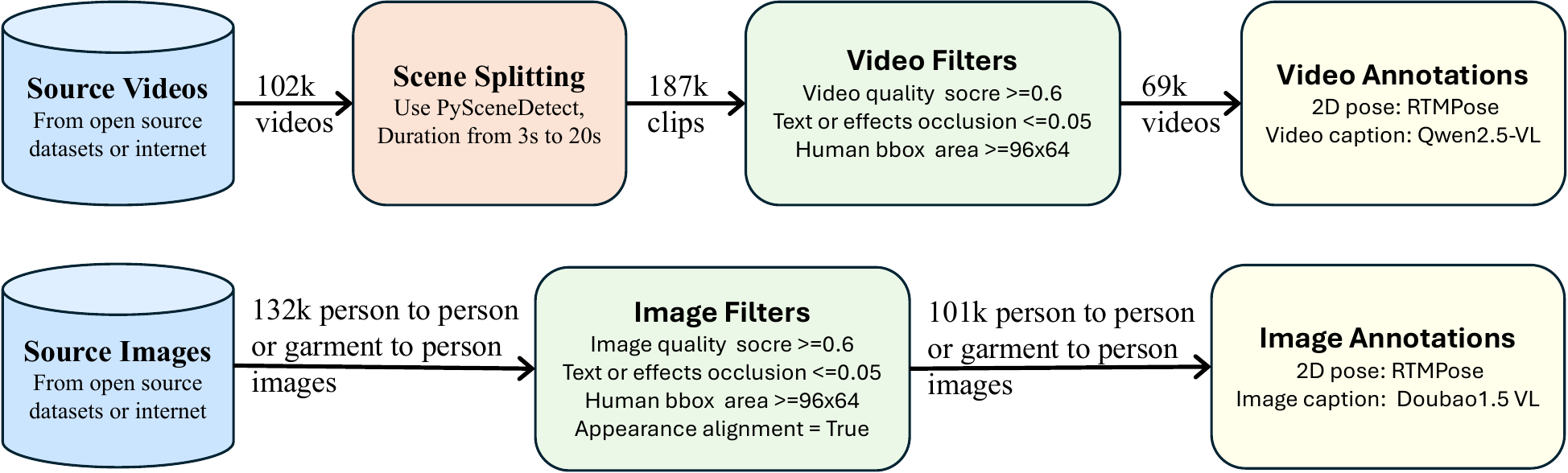}
    \captionof{figure}{Our curated and annotated pipeline for constructing human-centric video and paired image datasets. }
    \label{fig:supple_vid_pipeline}
    \vspace{2em}
}]

\subsection{Dataset Details}
\subsubsection{Video Dataset Construction }
To obtain high-quality video data for diverse scenarios, we first collected approximately 102K videos from public datasets \cite{li2024OpenHumanVid} and online sources. We then used PySceneDetect to segment these videos into shots, resulting in 187K video clips, each with a duration limited to between 3 and 20 seconds. Subsequently, we applied a series of filtering operators to remove invalid clips. Specifically, we filtered out low-quality videos based on VQA scores \cite{wu2023dover}, discarded nearly static videos according to motion strength \cite{dai2023animateanything}, and used video OCR (tesserocr) to detect and exclude clips with severe occlusion by identifying extensive text regions.
For annotation, we adopted RTMPose \cite{jiang2023rtmpose} as a robust and efficient pose representation. To extract textual descriptions from the raw videos, we employed the Qwen2.5-VL model \cite{Qwen2.5-VL} using a predefined template shown as Listing \ref{lst:VideoPrompt} that covers three aspects: environment, character appearance, and motion. In particular, the motion descriptions not only capture the overall movement of the person, but also provide detailed information about interactions with clothing and the environment.
During training, we randomly drop the appearance and environment descriptions to encourage the model to focus more on the motion descriptions. When generating try-on videos, the character appearance from the original video is replaced with that from the try-on image. 

\begin{lstlisting}[language=Python, caption={Video caption prompt}, label={lst:VideoPrompt},  breaklines=true, breakindent=0pt]]
prompt = f"""Describe input video  in the following json format:
{{"ENVIRONMENT": "Sentences describing the environment, video source and tags", 
"APPERANCE": "Sentences describing the character's apperance", 
"MOTION": "Sentences describing the character's actions, expressions and interaction with the environment. Do not add any action descriptions that are not present in the original description."}}
Only return in this format. Make sure that the output text can be directly parsed by a JSON parser. Do not add objects or character actions that do not exist in the original video description. Make sure that the output text should match the tone of video caption generation, therefore, phrases like "as described" should not appear."""
\end{lstlisting}

\subsubsection{Image Dataset Construction }
To construct a multi-frame consistent dataset for training multi-frame try-on models, in addition to sampling keyframes from videos, we also fully leveraged high-quality paired images with diverse viewpoints and poses collected from public websites and datasets. Specifically, after applying a series of filtering steps such as image quality assessment, appearance consistency check, and person size verification to the collected 1.3 million images, we obtained a clean set of 1.01 million images.
For image caption, since our model targets at synthesizing multi-frame virtual try-on results, we prepare elaborately designed prompt for every frame generation, which features detailed orientation description like front, back, side, and common identity and environment descriptions including clothing details as well as scenario background. Specifically, we caption every sampled frame to get appearance-unrelated information such as poseture and orientations, and then merge it with input garment information to get the frame-independent prompt. Subsequently, to ensure the consistency of prompt description, we also employ a VLM rewrite procedure   to eliminate potential conflicts introduced by the captioning model.


\subsection{Implementation Details Supplement}

\subsubsection{Keyframe Sampling Algorithm}
To facilitate reproducibility, we provide the pseudocode for the Keyframe Sampling strategy, as shown  Algorithm \ref{alg:KeyframeSampling}.
\begin{algorithm}[tb]
\caption{Keyframe Sampling}
\label{alg:KeyframeSampling}
\textbf{Input}: \\ 
Input video: $\mathbf{V_{in}}$\\ 
Anchor frame:  $f_{anchor}$ \\
\textbf{Parameter}: \\
Total number of frames in the input video: $N$  \\ 
Number of selected key frames: $K=2$ \\
Weight of the area ratio: $\lambda = 0.3$ \\
weight of score interval: $\alpha = 0.2$ \\
\textbf{Output}: \\
Key frame images: $\mathbf{f_{key}}$  
\begin{algorithmic}[1] 
\STATE Obtain skeletal joint direction vectors of anchor frame \\
$\mathbf{D_{anchor}}= compute\_joint\_direction(f_{anchor})$ 
\STATE Let $i=0$
\WHILE{$i < N$}
\STATE Obtain skeletal joint direction vectors of video frame\\
$\mathbf{D_{v}}[i]= compute\_joint\_direction(\mathbf{V_{in}}[i])$  \\
\STATE Compute motion similarity \\
$S_{m}[i] = \sum \mathbf{D_{anchor}} \cdot \mathbf{D_{v}}[i]$ \\
\STATE Compute area ratio \\
$S_r[i] = A_{subject} / A_{frame}$
\STATE Update final score \\
$S_{final}.update(index=i, score=S_{m}[i] + \lambda * S_r[i])$
\ENDWHILE
\STATE Calculate minimum score interval \\
$T_{s\_min} = \alpha  * get\_scores(S_{final}).mean $
\STATE $S_{final}= sort(S_{final}, score:descending) $
\STATE Initial key frame index \\
$\mathbf{Idx_{key}}[0] = get\_indexes(S_{final})[0]$
\STATE Let $i=N-1$
\WHILE{$i >=0$}
    \STATE $cur\_idx = get\_indexes(S_{final})[i]$ 
    \IF {$All(|(S_{final}[cur\_idx] - S_{final}[\mathbf{Idx_{key}}[:]]| >= T_{s\_min})$}
\STATE $\mathbf{Idx_{key}}.append(get\_indexes(S_{final})[i])$
    \ENDIF 
\ENDWHILE
\STATE $\mathbf{f_{key}}[:K] = \mathbf{V_{in}}[\mathbf{Idx_{key}}[:K]]$
\STATE \textbf{return} $\mathbf{f_{key}}$
\end{algorithmic}
\end{algorithm}

\subsubsection{Long-term Video Generation}
A common strategy for long video generation is to iteratively synthesize video segments such that the initial frames of the subsequent segment overlap with the final frames of the preceding segment. However, we observe that repeatedly passing the generated results of the previous segment through the decoder and encoder leads to the accumulation of errors over time, which results in noticeable degradation in the quality of the generated video as the sequence length increases. Therefore, we remove the decoder and encoder processes, and instead directly use the latent representation of the last frame from the previous video segment as the initial frame for the subsequent segment. By leveraging the pretrained model’s image-to-video generation capability for continuation, this approach significantly extends the duration before noticeable degradation occurs.

\subsection{More Results}\label{sec:MR}
\begin{figure*}
  \centering
    \captionsetup{type=figure}
    \includegraphics[width=0.9\textwidth]{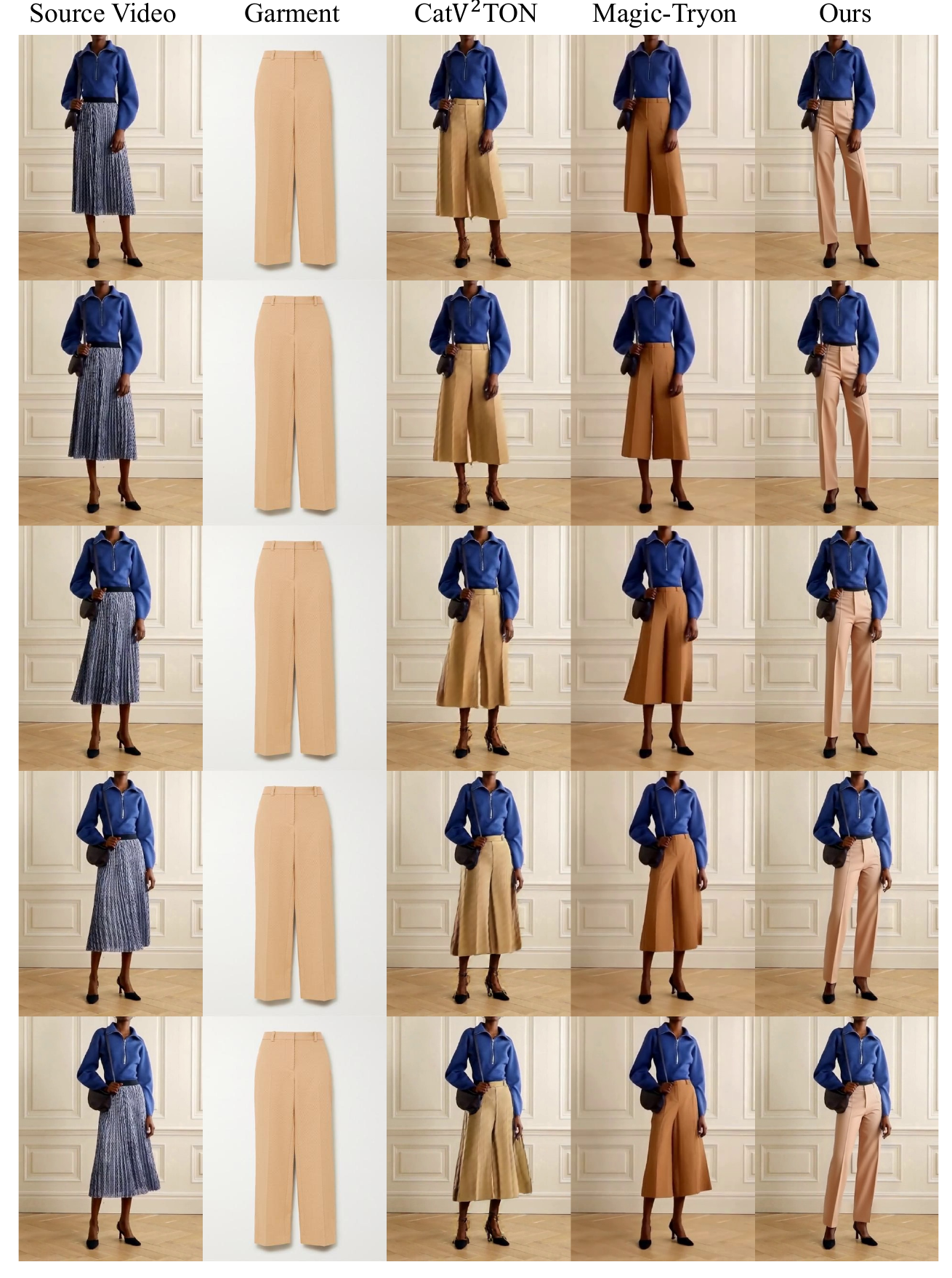}
  \caption{Additional qualitative comparison results on the ViViD dataset.Please zoom in for more details.} 
  \label{fig:supple4}
\vspace{-5mm}
\end{figure*}

\begin{figure*}
  \centering
    \captionsetup{type=figure}
    \includegraphics[width=0.9\textwidth]{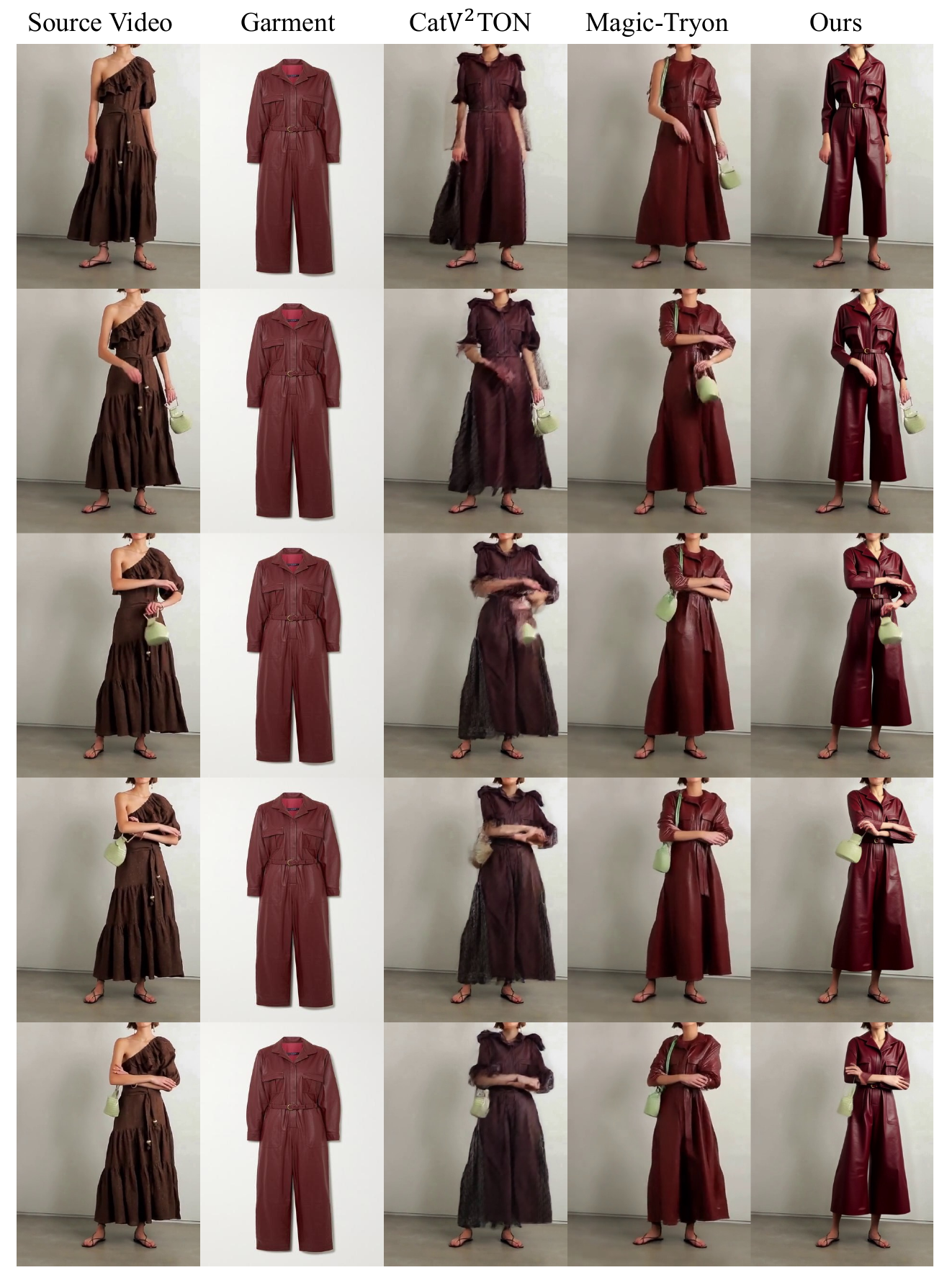}
  \caption{Additional qualitative comparison results on the ViViD dataset.Please zoom in for more details.} 
  \label{fig:supple6}
\vspace{-5mm}
\end{figure*}

\begin{figure*}
  \centering
    \captionsetup{type=figure}
    \includegraphics[width=0.85\textwidth]{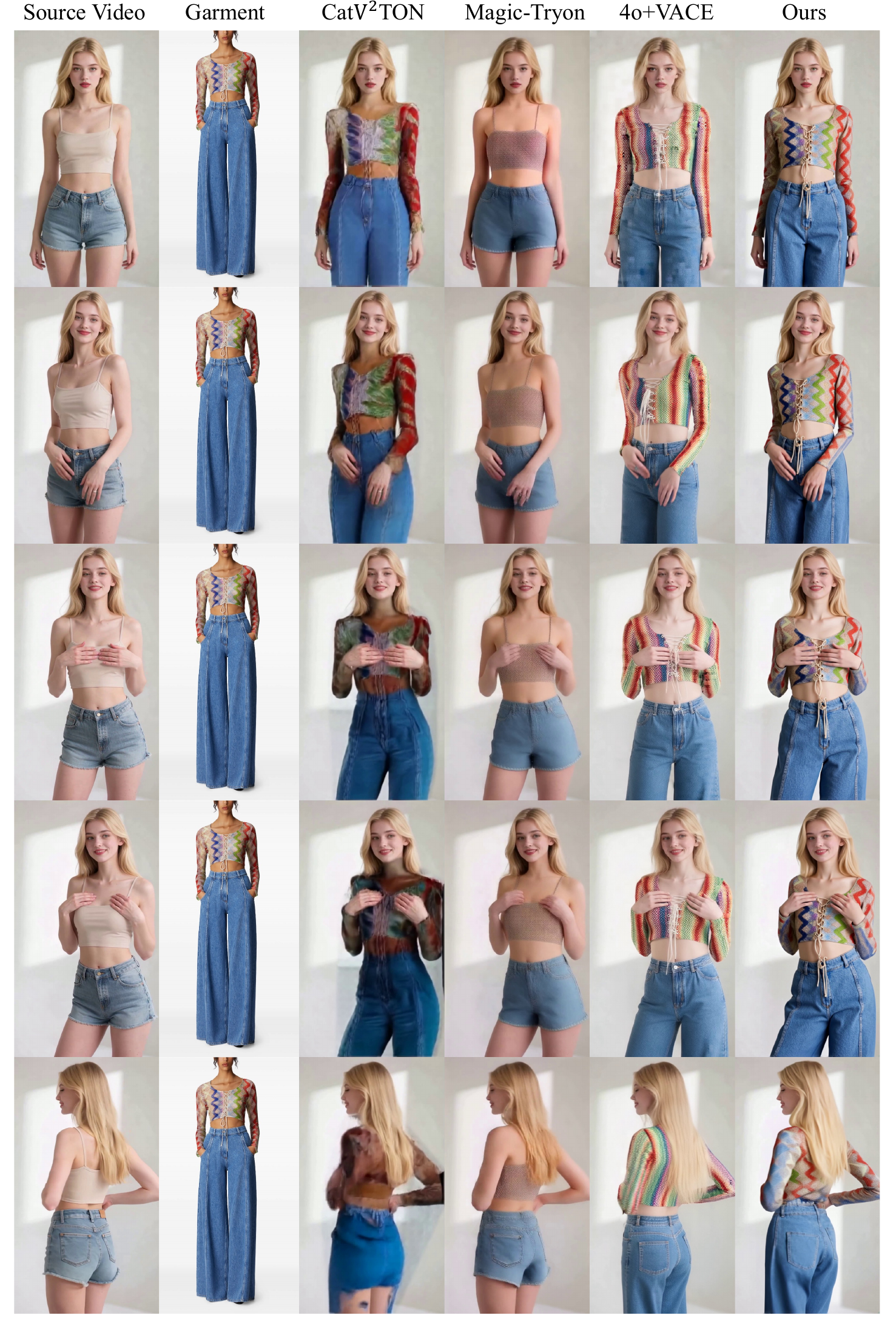}
  \caption{Additional qualitative comparison results on the WildTryOn dataset.Please zoom in for more details.} 
  \label{fig:supple1}
\vspace{-5mm}
\end{figure*}

\begin{figure*}
  \centering
    \captionsetup{type=figure}
    \includegraphics[width=0.85\textwidth]{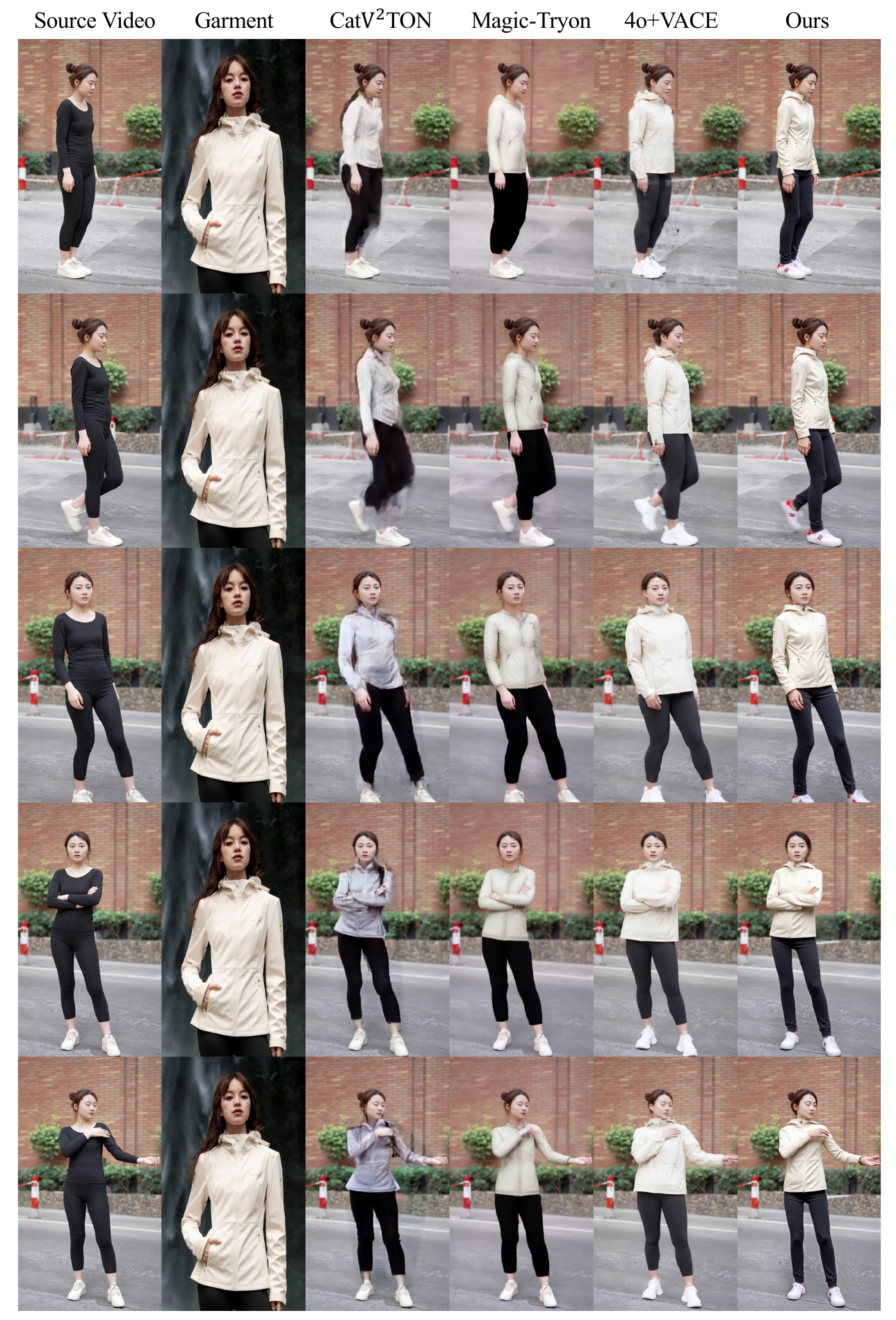}
  \caption{Additional qualitative comparison results on the WildTryOn dataset.Please zoom in for more details.} 
  \label{fig:supple2}
\vspace{-5mm}
\end{figure*}

\begin{figure*}[t]
  \centering
    \captionsetup{type=figure}
    \includegraphics[width=0.95\textwidth]{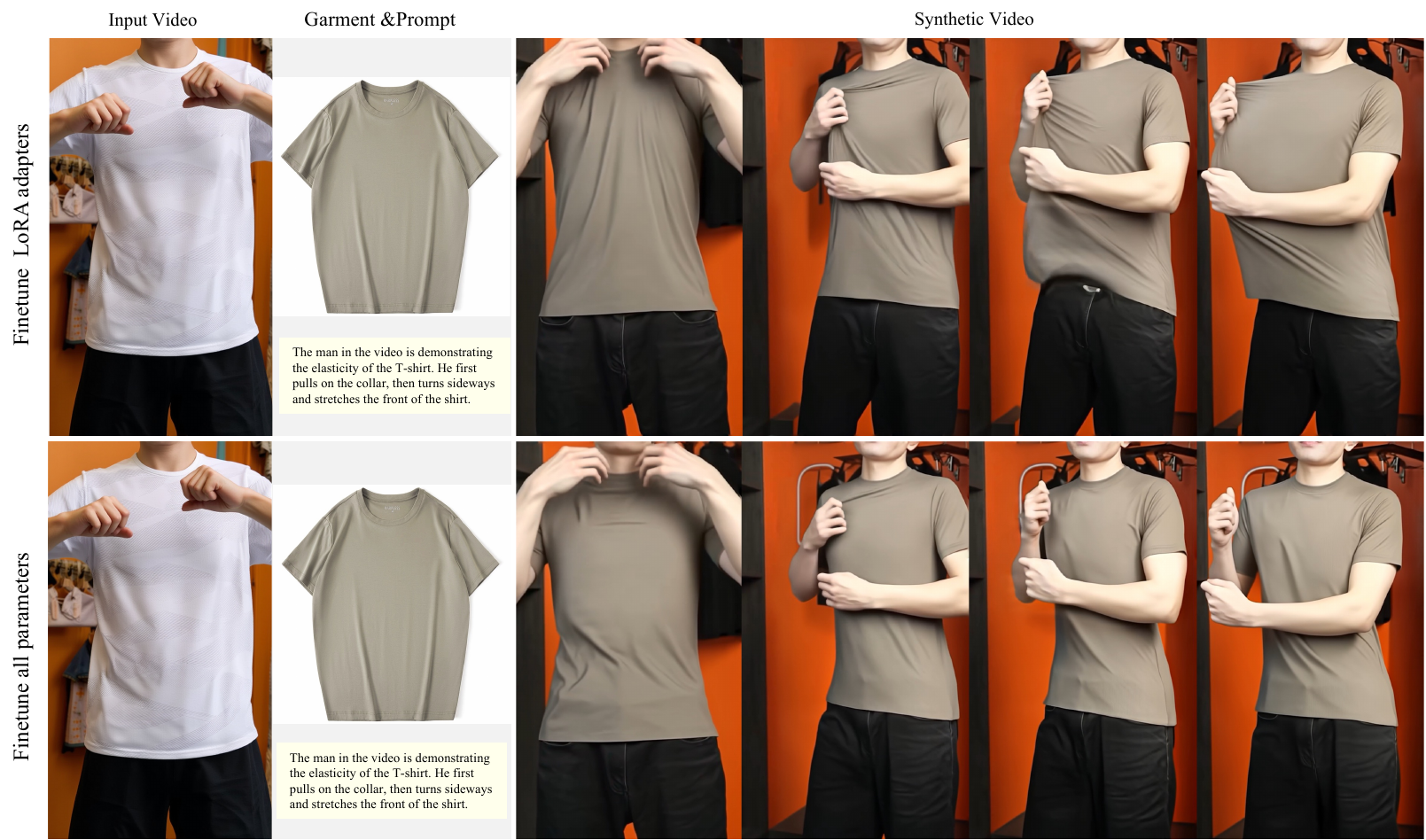}
  \caption{Additional ablation results: finetuning LoRA adapters vs. full model parameters.} 
  \label{fig:supple_lora}
\vspace{-5mm}
\end{figure*}

Fig.\ref{fig:supple4} and Fig.\ref{fig:supple6} show additional visual comparison results for our proposed DreamVVT network and the baseline methods on the ViViD datasets. While Fig.\ref{fig:supple1} and Fig.\ref{fig:supple2} show additional comparison results on the WildTryOn dataset.  Fig. \ref{fig:supple_lora} presents an additional ablation result, demonstrating that training LoRA adapters enables garment interaction more effectively than full parameter finetuning.  

%% file: main.bbl
\begin{thebibliography}{57}
\providecommand{\natexlab}[1]{#1}
\providecommand{\url}[1]{\texttt{#1}}
\expandafter\ifx\csname urlstyle\endcsname\relax
  \providecommand{\doi}[1]{doi: #1}\else
  \providecommand{\doi}{doi: \begingroup \urlstyle{rm}\Url}\fi

\bibitem[Bai et~al.(2023)Bai, Bai, Chu, Cui, Dang, Deng, Fan, Ge, Han, Huang, Hui, Ji, Li, Lin, Lin, Liu, Liu, Lu, Lu, Ma, Men, Ren, Ren, Tan, Tan, Tu, Wang, Wang, Wang, Wu, Xu, Xu, Yang, Yang, Yang, Yang, Yao, Yu, Yuan, Yuan, Zhang, Zhang, Zhang, Zhang, Zhou, Zhou, Zhou, and Zhu]{qwen}
Jinze Bai, Shuai Bai, Yunfei Chu, Zeyu Cui, Kai Dang, Xiaodong Deng, Yang Fan, Wenbin Ge, Yu Han, Fei Huang, Binyuan Hui, Luo Ji, Mei Li, Junyang Lin, Runji Lin, Dayiheng Liu, Gao Liu, Chengqiang Lu, Keming Lu, Jianxin Ma, Rui Men, Xingzhang Ren, Xuancheng Ren, Chuanqi Tan, Sinan Tan, Jianhong Tu, Peng Wang, Shijie Wang, Wei Wang, Shengguang Wu, Benfeng Xu, Jin Xu, An Yang, Hao Yang, Jian Yang, Shusheng Yang, Yang Yao, Bowen Yu, Hongyi Yuan, Zheng Yuan, Jianwei Zhang, Xingxuan Zhang, Yichang Zhang, Zhenru Zhang, Chang Zhou, Jingren Zhou, Xiaohuan Zhou, and Tianhang Zhu.
\newblock Qwen technical report.
\newblock \emph{arXiv preprint arXiv:2309.16609}, 2023.

\bibitem[Bai et~al.(2025)Bai, Chen, Liu, Wang, Ge, Song, Dang, Wang, Wang, Tang, Zhong, Zhu, Yang, Li, Wan, Wang, Ding, Fu, Xu, Ye, Zhang, Xie, Cheng, Zhang, Yang, Xu, and Lin]{Qwen2.5-VL}
Shuai Bai, Keqin Chen, Xuejing Liu, Jialin Wang, Wenbin Ge, Sibo Song, Kai Dang, Peng Wang, Shijie Wang, Jun Tang, Humen Zhong, Yuanzhi Zhu, Mingkun Yang, Zhaohai Li, Jianqiang Wan, Pengfei Wang, Wei Ding, Zheren Fu, Yiheng Xu, Jiabo Ye, Xi Zhang, Tianbao Xie, Zesen Cheng, Hang Zhang, Zhibo Yang, Haiyang Xu, and Junyang Lin.
\newblock Qwen2.5-vl technical report.
\newblock \emph{arXiv preprint arXiv:2502.13923}, 2025.

\bibitem[Blattmann et~al.(2023)Blattmann, Dockhorn, Kulal, Mendelevitch, Kilian, Lorenz, Levi, English, Voleti, Letts, et~al.]{blattmann2023stable}
Andreas Blattmann, Tim Dockhorn, Sumith Kulal, Daniel Mendelevitch, Maciej Kilian, Dominik Lorenz, Yam Levi, Zion English, Vikram Voleti, Adam Letts, et~al.
\newblock Stable video diffusion: Scaling latent video diffusion models to large datasets.
\newblock \emph{arXiv preprint arXiv:2311.15127}, 2023.

\bibitem[Carreira and Zisserman(2017)]{carreira2017quo}
Joao Carreira and Andrew Zisserman.
\newblock Quo vadis, action recognition? a new model and the kinetics dataset.
\newblock In \emph{proceedings of the IEEE Conference on Computer Vision and Pattern Recognition}, pages 6299--6308, 2017.

\bibitem[Choi et~al.(2021)Choi, Park, Lee, and Choo]{choi2021viton}
Seunghwan Choi, Sunghyun Park, Minsoo Lee, and Jaegul Choo.
\newblock Viton-hd: High-resolution virtual try-on via misalignment-aware normalization.
\newblock In \emph{Proceedings of the IEEE/CVF conference on computer vision and pattern recognition}, pages 14131--14140, 2021.

\bibitem[Chong et~al.(2024)Chong, Dong, Li, Zhang, Zhang, Zhang, Zhao, Jiang, and Liang]{chong2024catvton}
Zheng Chong, Xiao Dong, Haoxiang Li, Shiyue Zhang, Wenqing Zhang, Xujie Zhang, Hanqing Zhao, Dongmei Jiang, and Xiaodan Liang.
\newblock Catvton: Concatenation is all you need for virtual try-on with diffusion models.
\newblock \emph{arXiv preprint arXiv:2407.15886}, 2024.

\bibitem[Chong et~al.(2025)Chong, Zhang, Zhang, Zheng, Dong, Li, Wu, Jiang, and Liang]{chong2025catv2ton}
Zheng Chong, Wenqing Zhang, Shiyue Zhang, Jun Zheng, Xiao Dong, Haoxiang Li, Yiling Wu, Dongmei Jiang, and Xiaodan Liang.
\newblock Catv2ton: Taming diffusion transformers for vision-based virtual try-on with temporal concatenation.
\newblock \emph{arXiv preprint arXiv:2501.11325}, 2025.

\bibitem[Dai et~al.(2023)Dai, Zhang, Yao, Qiu, Zhu, Qin, and Wang]{dai2023animateanything}
Zuozhuo Dai, Zhenghao Zhang, Yao Yao, Bingxue Qiu, Siyu Zhu, Long Qin, and Weizhi Wang.
\newblock Animateanything: Fine-grained open domain image animation with motion guidance, 2023.

\bibitem[Dong et~al.(2019)Dong, Liang, Shen, Wu, Chen, and Yin]{dong2019fw}
Haoye Dong, Xiaodan Liang, Xiaohui Shen, Bowen Wu, Bing-Cheng Chen, and Jian Yin.
\newblock Fw-gan: Flow-navigated warping gan for video virtual try-on.
\newblock In \emph{Proceedings of the IEEE/CVF international conference on computer vision}, pages 1161--1170, 2019.

\bibitem[Esser et~al.(2024)Esser, Kulal, Blattmann, Entezari, M{\"u}ller, Saini, Levi, Lorenz, Sauer, Boesel, et~al.]{esser2024scaling}
Patrick Esser, Sumith Kulal, Andreas Blattmann, Rahim Entezari, Jonas M{\"u}ller, Harry Saini, Yam Levi, Dominik Lorenz, Axel Sauer, Frederic Boesel, et~al.
\newblock Scaling rectified flow transformers for high-resolution image synthesis.
\newblock In \emph{Forty-first international conference on machine learning}, 2024.

\bibitem[Fang et~al.(2024)Fang, Zhai, Su, Song, Zhu, Wang, Chen, Liu, Cao, and Zha]{fang2024vivid}
Zixun Fang, Wei Zhai, Aimin Su, Hongliang Song, Kai Zhu, Mao Wang, Yu Chen, Zhiheng Liu, Yang Cao, and Zheng-Jun Zha.
\newblock Vivid: Video virtual try-on using diffusion models.
\newblock \emph{arXiv preprint arXiv:2405.11794}, 2024.

\bibitem[Gao et~al.(2025)Gao, Gong, Guo, Hou, Lai, Li, Li, Lian, Liao, Liu, et~al.]{gao2025seedream}
Yu Gao, Lixue Gong, Qiushan Guo, Xiaoxia Hou, Zhichao Lai, Fanshi Li, Liang Li, Xiaochen Lian, Chao Liao, Liyang Liu, et~al.
\newblock Seedream 3.0 technical report.
\newblock \emph{arXiv preprint arXiv:2504.11346}, 2025.

\bibitem[Gou et~al.(2023)Gou, Sun, Zhang, Si, Qian, and Zhang]{gou2023taming}
Junhong Gou, Siyu Sun, Jianfu Zhang, Jianlou Si, Chen Qian, and Liqing Zhang.
\newblock Taming the power of diffusion models for high-quality virtual try-on with appearance flow.
\newblock In \emph{Proceedings of the 31st ACM International Conference on Multimedia}, pages 7599--7607, 2023.

\bibitem[Guo et~al.(2025)Guo, Zeng, Song, Zhang, Zhang, and Liu]{guo2025any2anytryon}
Hailong Guo, Bohan Zeng, Yiren Song, Wentao Zhang, Chuang Zhang, and Jiaming Liu.
\newblock Any2anytryon: Leveraging adaptive position embeddings for versatile virtual clothing tasks.
\newblock \emph{arXiv preprint arXiv:2501.15891}, 2025.

\bibitem[Hu et~al.(2022)Hu, Shen, Wallis, Allen-Zhu, Li, Wang, Wang, Chen, et~al.]{hu2022lora}
Edward~J Hu, Yelong Shen, Phillip Wallis, Zeyuan Allen-Zhu, Yuanzhi Li, Shean Wang, Lu Wang, Weizhu Chen, et~al.
\newblock Lora: Low-rank adaptation of large language models.
\newblock \emph{ICLR}, 1\penalty0 (2):\penalty0 3, 2022.

\bibitem[Hu(2024)]{hu2024animate}
Li Hu.
\newblock Animate anyone: Consistent and controllable image-to-video synthesis for character animation.
\newblock In \emph{Proceedings of the IEEE/CVF Conference on Computer Vision and Pattern Recognition}, pages 8153--8163, 2024.

\bibitem[Hu et~al.(2025{\natexlab{a}})Hu, Wang, Shen, Gao, Meng, Zhuo, Zhang, Zhang, and Bo]{hu2025animate}
Li Hu, Guangyuan Wang, Zhen Shen, Xin Gao, Dechao Meng, Lian Zhuo, Peng Zhang, Bang Zhang, and Liefeng Bo.
\newblock Animate anyone 2: High-fidelity character image animation with environment affordance.
\newblock \emph{arXiv preprint arXiv:2502.06145}, 2025{\natexlab{a}}.

\bibitem[Hu et~al.(2025{\natexlab{b}})Hu, Yu, Zhou, Liang, Zhou, Lin, and Lu]{hu2025hunyuancustom}
Teng Hu, Zhentao Yu, Zhengguang Zhou, Sen Liang, Yuan Zhou, Qin Lin, and Qinglin Lu.
\newblock Hunyuancustom: A multimodal-driven architecture for customized video generation.
\newblock \emph{arXiv preprint arXiv:2505.04512}, 2025{\natexlab{b}}.

\bibitem[Jiang et~al.(2025{\natexlab{a}})Jiang, Lin, Rong, Liang, Zhu, Yang, and Zhong]{jiang2025mobileportrait}
Jianwen Jiang, Gaojie Lin, Zhengkun Rong, Chao Liang, Yongming Zhu, Jiaqi Yang, and Tianyun Zhong.
\newblock Mobileportrait: Real-time one-shot neural head avatars on mobile devices.
\newblock In \emph{Proceedings of the Computer Vision and Pattern Recognition Conference}, pages 15920--15929, 2025{\natexlab{a}}.

\bibitem[Jiang et~al.(2023)Jiang, Lu, Zhang, Ma, Han, Lyu, Li, and Chen]{jiang2023rtmpose}
Tao Jiang, Peng Lu, Li Zhang, Ningsheng Ma, Rui Han, Chengqi Lyu, Yining Li, and Kai Chen.
\newblock Rtmpose: Real-time multi-person pose estimation based on mmpose.
\newblock \emph{arXiv preprint arXiv:2303.07399}, 2023.

\bibitem[Jiang et~al.(2025{\natexlab{b}})Jiang, Han, Mao, Zhang, Pan, and Liu]{vace}
Zeyinzi Jiang, Zhen Han, Chaojie Mao, Jingfeng Zhang, Yulin Pan, and Yu Liu.
\newblock Vace: All-in-one video creation and editing.
\newblock \emph{arXiv preprint arXiv:2503.07598}, 2025{\natexlab{b}}.

\bibitem[Karras et~al.(2022)Karras, Aittala, Aila, and Laine]{2022Elucidating}
Tero Karras, Miika Aittala, Timo Aila, and Samuli Laine.
\newblock Elucidating the design space of diffusion-based generative models.
\newblock 2022.

\bibitem[Kim et~al.(2024)Kim, Gu, Park, Park, and Choo]{kim2024stableviton}
Jeongho Kim, Guojung Gu, Minho Park, Sunghyun Park, and Jaegul Choo.
\newblock Stableviton: Learning semantic correspondence with latent diffusion model for virtual try-on.
\newblock In \emph{Proceedings of the IEEE/CVF conference on computer vision and pattern recognition}, pages 8176--8185, 2024.

\bibitem[Labs(2024)]{flux2024}
Black~Forest Labs.
\newblock Flux.
\newblock \url{https://github.com/black-forest-labs/flux}, 2024.

\bibitem[Li et~al.(2025{\natexlab{a}})Li, Zhong, Yu, Pan, Zhang, Yao, Han, and Mei]{li2025pursuing}
Dong Li, Wenqi Zhong, Wei Yu, Yingwei Pan, Dingwen Zhang, Ting Yao, Junwei Han, and Tao Mei.
\newblock Pursuing temporal-consistent video virtual try-on via dynamic pose interaction.
\newblock \emph{arXiv preprint arXiv:2505.16980}, 2025{\natexlab{a}}.

\bibitem[Li et~al.(2025{\natexlab{b}})Li, Zheng, Zhang, Chen, Luan, Ou, Zhao, Li, and Jiang]{li2025magictryon}
Guangyuan Li, Siming Zheng, Hao Zhang, Jinwei Chen, Junsheng Luan, Binkai Ou, Lei Zhao, Bo Li, and Peng-Tao Jiang.
\newblock Magictryon: Harnessing diffusion transformer for garment-preserving video virtual try-on, 2025{\natexlab{b}}.

\bibitem[Li et~al.(2024{\natexlab{a}})Li, Xu, Zhan, Mu, Li, Cheng, Chen, Chen, Ye, Wang, and Zhu]{li2024OpenHumanVid}
Hui Li, Mingwang Xu, Yun Zhan, Shan Mu, Jiaye Li, Kaihui Cheng, Yuxuan Chen, Tan Chen, Mao Ye, Jingdong Wang, and Siyu Zhu.
\newblock Openhumanvid: A large-scale high-quality dataset for enhancing human-centric video generation, 2024{\natexlab{a}}.

\bibitem[Li et~al.(2025{\natexlab{c}})Li, Jiang, Zhou, Liu, Chi, and Wang]{li2025realvvt}
Siqi Li, Zhengkai Jiang, Jiawei Zhou, Zhihong Liu, Xiaowei Chi, and Haoqian Wang.
\newblock Realvvt: Towards photorealistic video virtual try-on via spatio-temporal consistency.
\newblock \emph{arXiv preprint arXiv:2501.08682}, 2025{\natexlab{c}}.

\bibitem[Li et~al.(2024{\natexlab{b}})Li, Zhang, Lin, Xiong, Long, Deng, Zhang, Liu, Huang, Xiao, Chen, He, Li, Li, Zhang, Quan, Lu, Huang, Yuan, Zheng, Li, Zhang, Zhang, Chen, Liu, Fang, Wang, Xue, Tao, Zhu, Liu, Lin, Sun, Li, Wang, Chen, Hu, Xiao, Chen, Liu, Liu, Wang, Yang, Jiang, and Lu]{li2024hunyuandit}
Zhimin Li, Jianwei Zhang, Qin Lin, Jiangfeng Xiong, Yanxin Long, Xinchi Deng, Yingfang Zhang, Xingchao Liu, Minbin Huang, Zedong Xiao, Dayou Chen, Jiajun He, Jiahao Li, Wenyue Li, Chen Zhang, Rongwei Quan, Jianxiang Lu, Jiabin Huang, Xiaoyan Yuan, Xiaoxiao Zheng, Yixuan Li, Jihong Zhang, Chao Zhang, Meng Chen, Jie Liu, Zheng Fang, Weiyan Wang, Jinbao Xue, Yangyu Tao, Jianchen Zhu, Kai Liu, Sihuan Lin, Yifu Sun, Yun Li, Dongdong Wang, Mingtao Chen, Zhichao Hu, Xiao Xiao, Yan Chen, Yuhong Liu, Wei Liu, Di Wang, Yong Yang, Jie Jiang, and Qinglin Lu.
\newblock Hunyuan-dit: A powerful multi-resolution diffusion transformer with fine-grained chinese understanding, 2024{\natexlab{b}}.

\bibitem[Lin et~al.(2025{\natexlab{a}})Lin, Zhang, Zhao, Luo, Dong, Zeng, and Liang]{lin2025dreamfit}
Ente Lin, Xujie Zhang, Fuwei Zhao, Yuxuan Luo, Xin Dong, Long Zeng, and Xiaodan Liang.
\newblock Dreamfit: Garment-centric human generation via a lightweight anything-dressing encoder.
\newblock In \emph{Proceedings of the AAAI Conference on Artificial Intelligence}, pages 5218--5226, 2025{\natexlab{a}}.

\bibitem[Lin et~al.(2025{\natexlab{b}})Lin, Jiang, Yang, Zheng, and Liang]{lin2025omnihuman1}
Gaojie Lin, Jianwen Jiang, Jiaqi Yang, Zerong Zheng, and Chao Liang.
\newblock Omnihuman-1: Rethinking the scaling-up of one-stage conditioned human animation models.
\newblock \emph{arXiv preprint arXiv:2502.01061}, 2025{\natexlab{b}}.

\bibitem[Lin et~al.(2025{\natexlab{c}})Lin, Xia, Ren, Yang, Xiao, and Jiang]{lin2025diffusion}
Shanchuan Lin, Xin Xia, Yuxi Ren, Ceyuan Yang, Xuefeng Xiao, and Lu Jiang.
\newblock Diffusion adversarial post-training for one-step video generation.
\newblock \emph{arXiv preprint arXiv:2501.08316}, 2025{\natexlab{c}}.

\bibitem[Liu et~al.(2019)Liu, Piao, Min, Luo, Ma, and Gao]{liu2019liquid}
Wen Liu, Zhixin Piao, Jie Min, Wenhan Luo, Lin Ma, and Shenghua Gao.
\newblock Liquid warping gan: A unified framework for human motion imitation, appearance transfer and novel view synthesis.
\newblock In \emph{Proceedings of the IEEE/CVF international conference on computer vision}, pages 5904--5913, 2019.

\bibitem[Men et~al.(2024)Men, Yao, Cui, and Bo]{men2024mimo}
Yifang Men, Yuan Yao, Miaomiao Cui, and Liefeng Bo.
\newblock Mimo: Controllable character video synthesis with spatial decomposed modeling.
\newblock \emph{arXiv preprint arXiv:2409.16160}, 2024.

\bibitem[Morelli et~al.(2022)Morelli, Fincato, Cornia, Landi, Cesari, and Cucchiara]{morelli2022dress}
Davide Morelli, Matteo Fincato, Marcella Cornia, Federico Landi, Fabio Cesari, and Rita Cucchiara.
\newblock Dress code: High-resolution multi-category virtual try-on.
\newblock In \emph{Proceedings of the IEEE/CVF conference on computer vision and pattern recognition}, pages 2231--2235, 2022.

\bibitem[Morelli et~al.(2023)Morelli, Baldrati, Cartella, Cornia, Bertini, and Cucchiara]{morelli2023ladi}
Davide Morelli, Alberto Baldrati, Giuseppe Cartella, Marcella Cornia, Marco Bertini, and Rita Cucchiara.
\newblock Ladi-vton: Latent diffusion textual-inversion enhanced virtual try-on.
\newblock In \emph{Proceedings of the 31st ACM international conference on multimedia}, pages 8580--8589, 2023.

\bibitem[Peebles and Xie(2023)]{peebles2023scalable}
William Peebles and Saining Xie.
\newblock Scalable diffusion models with transformers.
\newblock In \emph{Proceedings of the IEEE/CVF international conference on computer vision}, pages 4195--4205, 2023.

\bibitem[Podell et~al.(2023)Podell, English, Lacey, Blattmann, Dockhorn, M{\"u}ller, Penna, and Rombach]{podell2023sdxl}
Dustin Podell, Zion English, Kyle Lacey, Andreas Blattmann, Tim Dockhorn, Jonas M{\"u}ller, Joe Penna, and Robin Rombach.
\newblock Sdxl: Improving latent diffusion models for high-resolution image synthesis.
\newblock \emph{arXiv preprint arXiv:2307.01952}, 2023.

\bibitem[Ren et~al.(2020)Ren, Yu, Chen, Li, and Li]{ren2020deep}
Yurui Ren, Xiaoming Yu, Junming Chen, Thomas~H Li, and Ge Li.
\newblock Deep image spatial transformation for person image generation.
\newblock In \emph{Proceedings of the IEEE/CVF conference on computer vision and pattern recognition}, pages 7690--7699, 2020.

\bibitem[Ren et~al.(2022)Ren, Fan, Li, Liu, and Li]{ren2022neural}
Yurui Ren, Xiaoqing Fan, Ge Li, Shan Liu, and Thomas~H Li.
\newblock Neural texture extraction and distribution for controllable person image synthesis.
\newblock In \emph{Proceedings of the IEEE/CVF conference on computer vision and pattern recognition}, pages 13535--13544, 2022.

\bibitem[Sun et~al.(2024)Sun, Cao, Wang, Tian, Zhang, Zhuo, Zhang, Bo, Zhou, Zhang, et~al.]{sun2024outfitanyone}
Ke Sun, Jian Cao, Qi Wang, Linrui Tian, Xindi Zhang, Lian Zhuo, Bang Zhang, Liefeng Bo, Wenbo Zhou, Weiming Zhang, et~al.
\newblock Outfitanyone: Ultra-high quality virtual try-on for any clothing and any person.
\newblock \emph{arXiv preprint arXiv:2407.16224}, 2024.

\bibitem[Team(2025)]{seed2025seed1_5vl}
ByteDance~Seed Team.
\newblock Seed1.5-vl technical report.
\newblock \emph{arXiv preprint arXiv:2505.07062}, 2025.

\bibitem[Wan et~al.(2025)Wan, Wang, Ai, Wen, Mao, Xie, Chen, Yu, Zhao, Yang, Zeng, Wang, Zhang, Zhou, Wang, Chen, Zhu, Zhao, Yan, Huang, Feng, Zhang, Li, Wu, Chu, Feng, Zhang, Sun, Fang, Wang, Gui, Weng, Shen, Lin, Wang, Wang, Zhou, Wang, Shen, Yu, Shi, Huang, Xu, Kou, Lv, Li, Liu, Wang, Zhang, Huang, Li, Wu, Liu, Pan, Zheng, Hong, Shi, Feng, Jiang, Han, Wu, and Liu]{wan2025}
Team Wan, Ang Wang, Baole Ai, Bin Wen, Chaojie Mao, Chen-Wei Xie, Di Chen, Feiwu Yu, Haiming Zhao, Jianxiao Yang, Jianyuan Zeng, Jiayu Wang, Jingfeng Zhang, Jingren Zhou, Jinkai Wang, Jixuan Chen, Kai Zhu, Kang Zhao, Keyu Yan, Lianghua Huang, Mengyang Feng, Ningyi Zhang, Pandeng Li, Pingyu Wu, Ruihang Chu, Ruili Feng, Shiwei Zhang, Siyang Sun, Tao Fang, Tianxing Wang, Tianyi Gui, Tingyu Weng, Tong Shen, Wei Lin, Wei Wang, Wei Wang, Wenmeng Zhou, Wente Wang, Wenting Shen, Wenyuan Yu, Xianzhong Shi, Xiaoming Huang, Xin Xu, Yan Kou, Yangyu Lv, Yifei Li, Yijing Liu, Yiming Wang, Yingya Zhang, Yitong Huang, Yong Li, You Wu, Yu Liu, Yulin Pan, Yun Zheng, Yuntao Hong, Yupeng Shi, Yutong Feng, Zeyinzi Jiang, Zhen Han, Zhi-Fan Wu, and Ziyu Liu.
\newblock Wan: Open and advanced large-scale video generative models.
\newblock \emph{arXiv preprint arXiv:2503.20314}, 2025.

\bibitem[Wang et~al.(2025)Wang, Zhang, Di, Zhang, and Zuo]{wang2025mv}
Haoyu Wang, Zhilu Zhang, Donglin Di, Shiliang Zhang, and Wangmeng Zuo.
\newblock Mv-vton: Multi-view virtual try-on with diffusion models.
\newblock In \emph{Proceedings of the AAAI Conference on Artificial Intelligence}, pages 7682--7690, 2025.

\bibitem[Wang et~al.(2024)Wang, Zhang, Gao, Wang, Zhou, Zhang, Yan, and Sang]{wang2024unianimate}
Xiang Wang, Shiwei Zhang, Changxin Gao, Jiayu Wang, Xiaoqiang Zhou, Yingya Zhang, Luxin Yan, and Nong Sang.
\newblock Unianimate: Taming unified video diffusion models for consistent human image animation.
\newblock \emph{arXiv preprint arXiv:2406.01188}, 2024.

\bibitem[Wang et~al.(2004)Wang, Bovik, Sheikh, and Simoncelli]{Wang2004SSIM}
Zhou Wang, Alan~Conrad Bovik, Hamid~Rahim Sheikh, and Eero~P Simoncelli.
\newblock Image quality assessment: from error visibility to structural similarity.
\newblock \emph{IEEE transactions on image processing}, 13\penalty0 (4):\penalty0 600--612, 2004.

\bibitem[Wu et~al.(2023)Wu, Zhang, Liao, Chen, Hou, Wang, Sun, Yan, and Lin]{wu2023dover}
Haoning Wu, Erli Zhang, Liang Liao, Chaofeng Chen, Jingwen~Hou Hou, Annan Wang, Wenxiu~Sun Sun, Qiong Yan, and Weisi Lin.
\newblock Exploring video quality assessment on user generated contents from aesthetic and technical perspectives.
\newblock In \emph{International Conference on Computer Vision (ICCV)}, 2023.

\bibitem[Xie et~al.(2017)Xie, Girshick, Doll{\'a}r, Tu, and He]{xie2017aggregated}
Saining Xie, Ross Girshick, Piotr Doll{\'a}r, Zhuowen Tu, and Kaiming He.
\newblock Aggregated residual transformations for deep neural networks.
\newblock In \emph{Proceedings of the IEEE conference on computer vision and pattern recognition}, pages 1492--1500, 2017.

\bibitem[Xie et~al.(2023)Xie, Huang, Dong, Zhao, Dong, Zhang, Zhu, and Liang]{xie2023gp}
Zhenyu Xie, Zaiyu Huang, Xin Dong, Fuwei Zhao, Haoye Dong, Xijin Zhang, Feida Zhu, and Xiaodan Liang.
\newblock Gp-vton: Towards general purpose virtual try-on via collaborative local-flow global-parsing learning.
\newblock In \emph{Proceedings of the IEEE/CVF conference on computer vision and pattern recognition}, pages 23550--23559, 2023.

\bibitem[Xu et~al.(2025)Xu, Gu, Chen, and Chen]{xu2025ootdiffusion}
Yuhao Xu, Tao Gu, Weifeng Chen, and Arlene Chen.
\newblock Ootdiffusion: Outfitting fusion based latent diffusion for controllable virtual try-on.
\newblock In \emph{Proceedings of the AAAI Conference on Artificial Intelligence}, pages 8996--9004, 2025.

\bibitem[Xu et~al.(2024)Xu, Chen, Wang, Xing, Zhai, Sang, Lan, Xiao, and Gao]{xu2024tunnel}
Zhengze Xu, Mengting Chen, Zhao Wang, Linyu Xing, Zhonghua Zhai, Nong Sang, Jinsong Lan, Shuai Xiao, and Changxin Gao.
\newblock Tunnel try-on: Excavating spatial-temporal tunnels for high-quality virtual try-on in videos.
\newblock In \emph{Proceedings of the 32nd ACM International Conference on Multimedia}, pages 3199--3208, 2024.

\bibitem[Ye et~al.(2025)Ye, He, Liu, Wang, Wang, Wan, Zhang, Gai, Chen, and Luo]{ye2025unic}
Zixuan Ye, Xuanhua He, Quande Liu, Qiulin Wang, Xintao Wang, Pengfei Wan, Di Zhang, Kun Gai, Qifeng Chen, and Wenhan Luo.
\newblock Unic: Unified in-context video editing.
\newblock \emph{arXiv preprint arXiv:2506.04216}, 2025.

\bibitem[Zhang et~al.(2018)Zhang, Isola, Efros, Shechtman, and Wang]{zhang2018unreasonable}
Richard Zhang, Phillip Isola, Alexei~A Efros, Eli Shechtman, and Oliver Wang.
\newblock The unreasonable effectiveness of deep features as a perceptual metric.
\newblock In \emph{Proceedings of the IEEE conference on computer vision and pattern recognition}, pages 586--595, 2018.

\bibitem[Zhang et~al.(2024{\natexlab{a}})Zhang, Song, Zhan, Chang, Zeng, Chen, Luo, and Liu]{zhang2024boow}
Xuanpu Zhang, Dan Song, Pengxin Zhan, Tianyu Chang, Jianhao Zeng, Qingguo Chen, Weihua Luo, and Anan Liu.
\newblock Boow-vton: Boosting in-the-wild virtual try-on via mask-free pseudo data training.
\newblock \emph{arXiv preprint arXiv:2408.06047}, 2024{\natexlab{a}}.

\bibitem[Zhang et~al.(2024{\natexlab{b}})Zhang, Gu, Wang, Wang, Cheng, Zhu, and Zou]{zhang2024mimicmotion}
Yuang Zhang, Jiaxi Gu, Li-Wen Wang, Han Wang, Junqi Cheng, Yuefeng Zhu, and Fangyuan Zou.
\newblock Mimicmotion: High-quality human motion video generation with confidence-aware pose guidance.
\newblock \emph{arXiv preprint arXiv:2406.19680}, 2024{\natexlab{b}}.

\bibitem[Zheng et~al.(2024{\natexlab{a}})Zheng, Wang, Zhao, Zhang, and Liang]{zheng2024dynamic}
Jun Zheng, Jing Wang, Fuwei Zhao, Xujie Zhang, and Xiaodan Liang.
\newblock Dynamic try-on: Taming video virtual try-on with dynamic attention mechanism.
\newblock \emph{arXiv preprint arXiv:2412.09822}, 2024{\natexlab{a}}.

\bibitem[Zheng et~al.(2024{\natexlab{b}})Zheng, Gao, Fan, Liu, Laaksonen, Ouyang, and Sebe]{zheng2024birefnet}
Peng Zheng, Dehong Gao, Deng-Ping Fan, Li Liu, Jorma Laaksonen, Wanli Ouyang, and Nicu Sebe.
\newblock Bilateral reference for high-resolution dichotomous image segmentation.
\newblock \emph{CAAI Artificial Intelligence Research}, 3:\penalty0 9150038, 2024{\natexlab{b}}.

\end{thebibliography}
